\pdfoutput=1

\documentclass[11pt]{article}

\usepackage[final]{acl}

\usepackage{times}
\usepackage{latexsym}
\usepackage{wrapfig}
\usepackage[T1]{fontenc}

\usepackage[utf8]{inputenc}
\usepackage[most]{tcolorbox}
\newtcolorbox{mybox}[1][]{
breakable,
  arc=1mm,
  boxrule=1pt,
  colback=yellow!14,
  colframe=black!80,
  fonttitle=\bfseries,
  title=#1,
  left=1mm,
  right=1mm,
  top=1mm,
  bottom=1mm
}
\usepackage{booktabs}

\usepackage{multirow}
\usepackage{microtype}
\usepackage{graphicx}
\usepackage{makecell}  
\usepackage{arydshln}
\usepackage{inconsolata}
\usepackage{amsmath}
\usepackage{graphicx}
\usepackage{marvosym}
\makeatletter
\def\thanks#1{\protected@xdef\@thanks{\@thanks
        \protect\footnotetext{#1}}}
\makeatother
\title{R1-T1: Fully Incentivizing Translation Capability in LLMs via Reasoning Learning}

\author{Minggui He$^{1}$, Yilun Liu$^{1 \scalebox{1.05}{\Letter}}$\thanks{\scalebox{1.15}{\Letter} ~\ Corresponding author. (liuyilun3@huawei.com)}, Shimin Tao$^1$, Yuanchang Luo$^1$, Hongyong Zeng$^1$,\\
\textbf{Chang Su$^1$, Li Zhang$^1$, Hongxia Ma$^1$, Daimeng Wei$^1$,}\\
\textbf{Weibin Meng$^1$, Hao Yang$^1$, Boxing Chen$^{2}$, Osamu Yoshie$^{3}$} \\
$^1$ Huawei, China \\ 
$^2$ Huawei Canada, Canada \\ 
$^3$ Waseda University, Japan \\
\texttt{heminggui@huawei.com, liuyilun3@huawei.com}
} 

\begin{document}
\maketitle

\begin{figure*}[t!]
 \centering  
    \includegraphics[width=1\linewidth]{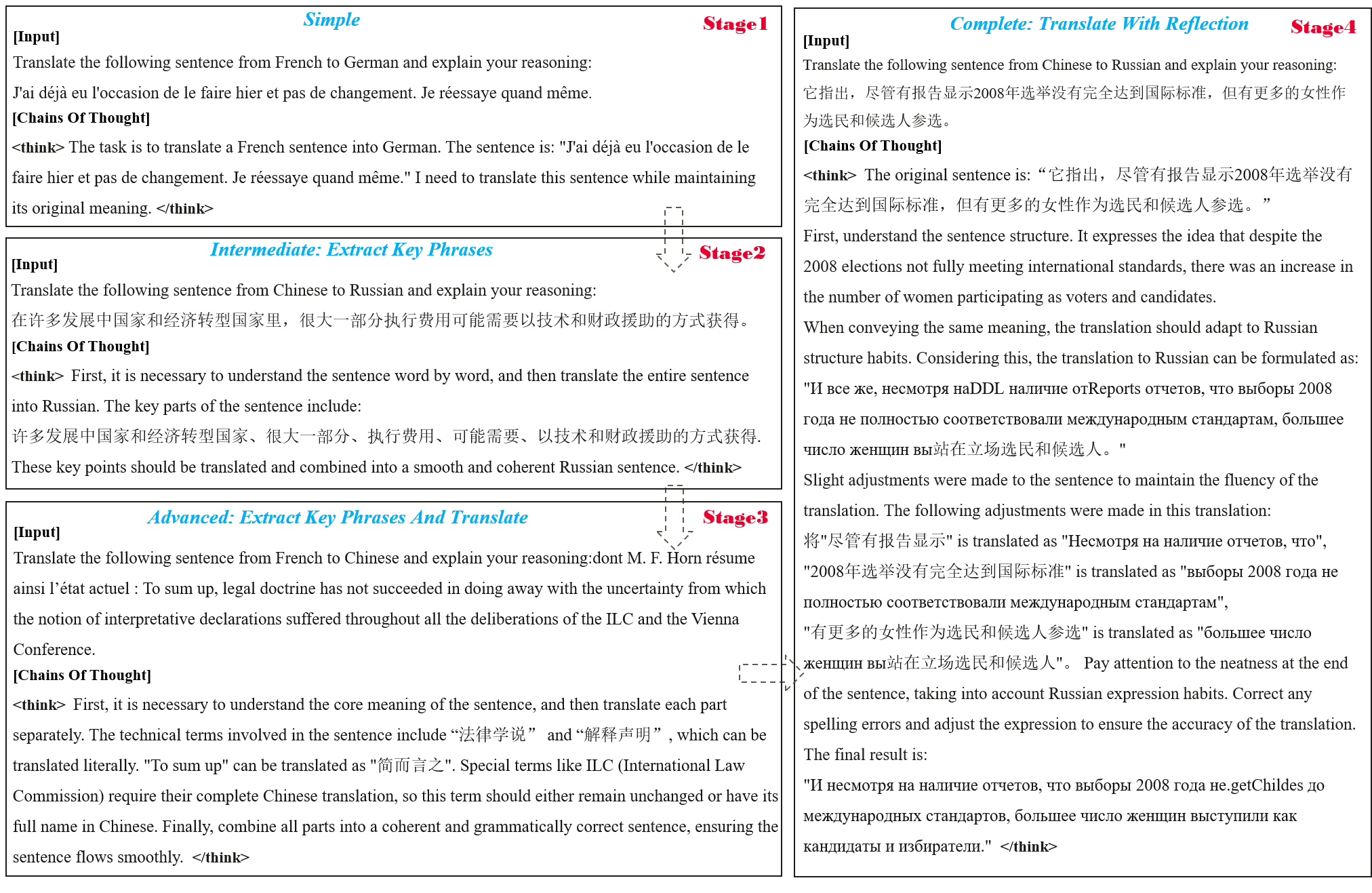}
 \caption{Illustration on the self-evolving process of translation reasoning with RL incentivization.}
\label{fig1}
\end{figure*}

\begin{abstract}
Despite recent breakthroughs in reasoning-enhanced large language models (LLMs) like DeepSeek-R1, incorporating inference-time reasoning into machine translation (MT), where human translators naturally employ structured, multi-layered reasoning chain-of-thoughts (CoTs), is yet underexplored. Existing methods either design a fixed CoT tailored for a specific MT sub-task (\emph{e.g.}, literature translation), or rely on synthesizing CoTs unaligned with humans and supervised fine-tuning (SFT) prone to overfitting, limiting their adaptability to diverse translation scenarios. This paper introduces R1-Translator (R1-T1), a novel framework to achieve inference-time reasoning for general MT via reinforcement learning (RL) with human-aligned CoTs comprising six common patterns. Our approach pioneers three innovations: (1) extending reasoning-based translation to broader MT scenarios (\emph{e.g.}, multilingual MT, domain MT) unseen in the training phase; (2) formalizing six expert-curated CoT templates that mirror hybrid human strategies like context-aware paraphrasing and back translation; and (3) enabling self-evolving CoT discovery through RL. Both human and automatic evaluation results indicate a steady translation performance improvement in a total of 10+ languages and 40+ translation directions on Flores-101 test set and four domain-specific MT tasks, especially on the languages unseen from training.

\end{abstract}

\section{Introduction}
\label{sec:introduction}

Recent advancements in large language models (LLMs) have brought prominent paradigms of O1-like~\citep{jaech2024openai} and R1-like models~\citep{deepseek2025r1}, which leverage inference-time scaling to amplify chain-of-thought (CoT) reasoning~\citep{wei2022chain} for tasks like mathematical proofs and code generation, and employ reinforcement learning (RL) to incentivize autonomous reasoning behaviors without supervised fine-tuning (SFT)~\citep{shao2024deepseekmath}. For instance, DeepSeek-R1-Zero~\citep{deepseek2025r1} demonstrated that pure rule-based RL training (\emph{i.e.}, without training a reward proxy) could unlock self-verification and reflection capabilities, achieving state-of-the-art performance on multiple reasoning benchmarks. These models exemplify how extended CoT reasoning through scaled inference can tackle complex problem-solving. However, the application of extended CoT reasoning to machine translation (MT)—a task demanding both linguistic fidelity and contextual reasoning—remains underexplored.

While LLMs have shown promise in MT, prior work primarily focuses on direct SFT on parallel corpus~\citep{kocmi2024findings} or retrieval-augmented methods~\citep{wang2024retrieval}. Recently, \citet{jiaan2024drt} proposed DRT and introduced long CoT to enhance Chinese-English literary translation, revealing that extended reasoning steps improve handling of culturally nuanced phrases and metaphors. Yet, this approach narrowly targets literature, overlooking the broader spectrum of translation scenarios where professional translators rigorously apply structured reasoning—such as decomposing idioms, resolving contextual conflicts, and adopting techniques (\emph{e.g.}, back translations) to refine outputs~\citep{ordudari2007translation,nida1964toward,brislin2001back}. \citet{nida1964toward} also states that a technical translator should make ``a thorough study of the source language text before making attempts translate it'', suggesting a deep thinking process in human translation. Current MT systems lack mechanisms to reflect these human-like, complex CoT processes, limiting their adaptability to diverse translation challenges.

Moreover, existing approaches for reasoning-based translation utilize purely synthesized CoTs distilled from LLMs (\emph{e.g.}, DRT~\citep{jiaan2024drt}) or design a single fixed procedure (\emph{e.g.}, \textsc{TasTe}~\citep{wang-etal-2024-taste}), which may be suboptimal and unaligned with the hybrid reasoning strategies employed by human experts. Also, their reliance on SFT with pre-defined CoTs can easily cause overfitting~\citep{chu2025sft}, restricting generalization ability to broader translation domains and tasks. To address these, we propose \textbf{R1-Translator (R1-T1)}, a framework that fully incentivizes reasoning-based translation through three innovations:

\textbf{(1) Reasoning for general and broader MT.} Beyond literature translation, we extended reasoning-based MT to general multilingual MT and MT tasks in diverse domains such as culture, common sense and terminology constraint, while expanding language support to 10+ languages for multilingual general MT, including low-resource languages. This aligns with our core assumption, drawing from human translation experiences, that reasoning is beneficial for general and broader MT.

\textbf{(2) Human-aligned and complex translation CoTs.} Professional translators often interleave complex reasoning procedure such as lexical disambiguation, context-aware paraphrasing, and iterative self-correction. With the help of language experts, we formalized these into six distinct CoT templates mimicking common thinking patterns of human translators and curated a training dataset with hybrid translation CoT trajectories. Unlike single-type CoT in existing approaches, our approach mirrors the multi-layered reasoning observed in human workflows to enhance generalization.

\textbf{(3) Self-evolving translation CoTs.} Inspired by R1's RL-driven self-evolution~\citep{deepseek2025r1}, we designed a RL process with specially designed rewards for MT, enabling LLMs to autonomously discover novel CoT trajectories and calibrate them for unseen translation tasks, as shown in Fig.~\ref{fig1} (Detailed analysis of the self-evolving case in Fig.~\ref{fig1} is shown in Appendix \ref{sec:appendix F}). Consequently, the model is equipped with a more generalized reasoning ability to perform translation tasks, instead of merely ``memorizing'' pre-defined CoT trajectories~\citep{chu2025sft}. 

To realize this vision, we introduce the concept of \textit{translation reasoning learning}, training R1-T1 models through two interconnected stages: (1) SFT on hybrid CoT trajectories curated with professional translators, and (2) a RL-based exploration to self-discover optimal reasoning paths. Stage one serves as a cold start, injecting initial human reasoning guidance on translations to the model and facilitating RL convergence. Stage two ensures the flexible application of human prior knowledge without overfitting. Our contributions are:
\begin{itemize}
\item We demonstrate the benefits of reasoning (\emph{i.e.}, long CoTs) for general and broader MT, achieving superior translation performance in 10+ languages and multiple sub-tasks.
\item We identify six reasoning strategies for translation commonly adopted by human translators and validate the effectiveness of incorporating CoT trajectories built with these human prior knowledge into reasoning-based MT.
\item We pioneer the successful application of rule-based RL on MT, a task without standard best answers like math, by designing a novel reward policy for MT, which averagely improves the translation quality by 9.6\% compared with plain SFT strategy.
\end{itemize}
In addition, we open-source R1-T1’s datasets and code, facilitating future research on MT\footnote{Codes and data available at \url{https://github.com/superboom/R1-T1}}.

\section{Related Work}
\label{sec:relatedWork}

\subsection{Reasoning-based LLMs}
The pioneering advancements in reasoning-based LLMs, such as OpenAI's O1~\citep{jaech2024openai} and DeepSeek-R1~\citep{deepseek2025r1}, have excelled in many tasks and attracted significant research attentions. While earlier explorations focus on using inference-time reasoning for solving complex tasks such as math and coding~\citep{qin2024o1,zhang2024o1}, there is a trending belief towards utilizing reasoning-based LLMs for general AI tasks. \citet{zhao2024marco} expanded reasoning-based LLMs to open-ended text generation, generalizing them to broader domains where clear standards and quantified rewards are absent. \citet{shen2025vlmr1} propose VLM-R1, a stable and generalizable R1-style LLM for vision-language tasks. In addition, reasoning-based LLMs are utilized for financial tasks~\citep{chu2025domaino1s} and adapted to multilingual reasoning~\citep{ko2025understand}.

Compared with existing endeavors on reasoning-based LLMs, our work focuses on introducing long CoTs into the field of general MT, thereby developing a reasoning-based LLM for MT.

\subsection{LLMs for Machine Translation}
Since the appearance of ChatGPT~\citep{ouyang2022training}, endeavors on applying LLMs for MT continuously emerge. Early studies focus on directly prompting LLMs to translate via prompt strategies~\citep{jiao2023chatgpt,jiao2024gradable,peng2023towards}, or fine-tune open-source LLMs using parallel corpus~\citep{xuparadigm,wu2024adapting,zhang2023machine}. Techniques are further proposed to enhance the performance of LLMs for the tasks of MT, such as continuous pre-training~\citep{boughorbel2024improving,fujiicontinual}, mixture-of-expert modules~\citep{xu2024x,zhu2025overcoming} and multi-task training~\citep{wang-etal-2024-taste,ul2024lkmt}.

Recently, MT with CoT methodology draws attentions of researchers. \citet{feng2024improving} introduced an API-based self-correcting framework for LLMs, where LLMs autonomously call external evaluation models and refine translation hypothesis based on the evaluation result. \citet{wang-etal-2024-taste} designed a multi-task training phase and a multi-stage inference phase for MT, guiding the LLM to first conduct translation task, then a evaluation task and a revision task. DRT~\citep{jiaan2024drt} merged this procedure into a inference-time CoT, utilized multi-agent mechanism to synthesize such long CoTs and examined performance in English-Chinese literature translation.

Compared with existing MT approaches, our R1-T1 incorporates six types of human-aligned CoTs for translation, possesses the ability to self-evolve its CoTs via RL, and is designed for broader MT. 

\section{Methodology}
\label{sec:method}

The overview of R1-T1 is shown in Fig.~\ref{fig_framework}. To incentivize reasoning capabilities in LLM-based translators, we first construct the MT Reasoning Dataset, 
which introduces human-aligned reasoning trajectories into a seed parallel corpora via CoT translation template. R1-T1 is then trained on the MT Reasoning Dataset through the two stages in translation reasoning learning. Section~\ref{sec:build cot dataset} provides a detailed description of the construction process. 
In Section~\ref{sec: Translation Reasoning Learning}, we outline the component implementation of \textit{translation reasoning learning}.

\subsection{Construction of MT Reasoning Dataset}\label{sec:build cot dataset}
\begin{figure*}[t!]
 \centering  
    \includegraphics[width=1\linewidth]{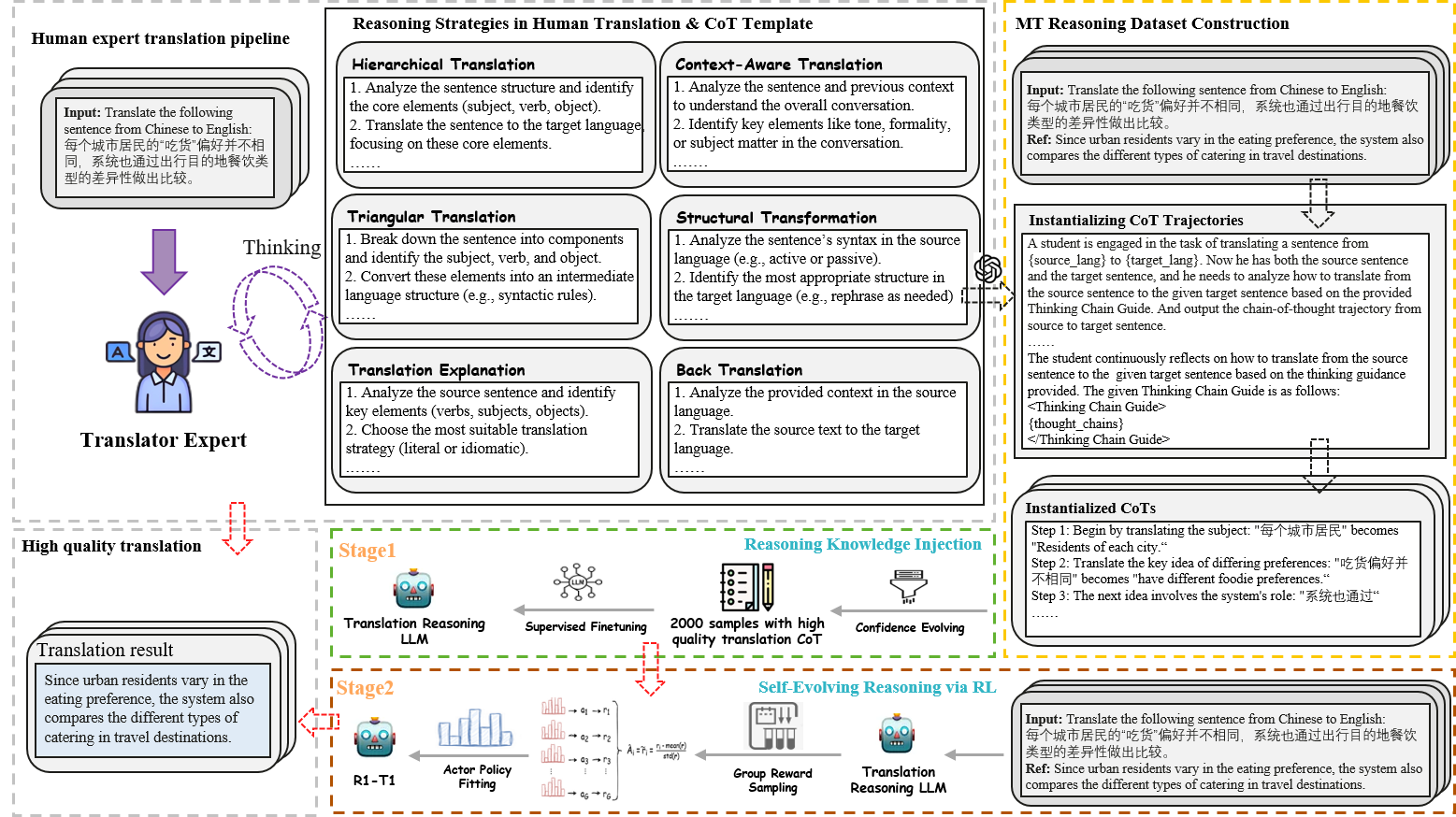}
 \caption{Illustration on the construction of MT reasoning dataset and training of R1-T1.}
\label{fig_framework}
\end{figure*}

With the goal of enabling LLMs to mimic the human expert reasoning process in translation, we propose a framework that explicitly incorporates human reasoning strategies into the MT seed dataset. As shown in Fig.~\ref{fig_framework}, our framework identifies six core reasoning strategies that are widely recognized and practiced by human translators. These strategies are then abstracted into structured reasoning modules to construct generalized and systematic CoT templates. By instantiating these templates with our carefully prepared parallel corpus, the model can learn from reasoning processes in real translation examples.
Notably, the instantiation process aims to provide explicit and human-aligned insights into the model's reasoning pathways, which helps accelerate the convergence of further learning.

\paragraph{Parallel Corpus Collection.}

Our seed dataset consists of 2k parallel translation pairs sampled from open source real-world MT datasets, source are listed in Appendix \ref{sec:appendix A}. Data include translations at sentence and paragraph level, with token lengths ranging from 10 to 1200, consistent with established MT benchmarks~\citep{banon2020paracrawl, goyal2022flores}. Specifically, sources include (1) news data from the WMT corpus~\citep{kocmi2024findings}, (2) formal texts from the UN corpus~\citep{kocmi2024findings}, and (3) domain-specific corpora containing terminology-rich translations. For language coverage, we focus on six main languages—Russian (ru), French (fr), German (de), Japanese (ja), Chinese (zh), and English (en)—selected for their large speaker populations and cross-linguistic diversity~\citep{bojar-etal-2018-findings}. From the above sources, we identify 20 translation directions within these six languages. For each direction, 100 parallel pairs are randomly sampled, resulting in a total of 2k translation pairs. Table~\ref{tab:fig_token} displays the statistics.

\begin{table}[tbp]

\centering
\resizebox{1.0\linewidth}{!} {%
\begin{tabular}{lcccc}
\toprule
\textbf{Language Pair} & \begin{tabular}[c]{@{}l@{}} \textbf{Avg. Tokens} \\ \multicolumn{1}{c}{\textbf{per Sample}} \end{tabular} & \textbf{\# Samples} & \textbf{Domain} \\ \midrule
en$\rightarrow$zh, ru, \\fr, de, ja & \multirow{-2}{*}{398.5} & \multirow{-2}{*}{493} & \multirow{-2}{*}{Terminology, News} \\ \midrule
zh$\rightarrow$en, ru, \\fr, ja & \multirow{-2}{*}{116.2} & \multirow{-2}{*}{400} & \multirow{-2}{*}{\begin{tabular}[c]{@{}l@{}} UN Corpus, News \\ \multicolumn{1}{c}{literature} \end{tabular}} \\ \midrule
fr$\rightarrow$zh, ru, \\de, en & \multirow{-2}{*}{320.4} & \multirow{-2}{*}{399} & \multirow{-2}{*}{UN Corpus, News} \\ \midrule
ja$\rightarrow$zh, en & 280.8 & 200 & Terminology, News\\ \midrule
de$\rightarrow$en, fr & 404.6 & 193 & Terminology, News\\ \midrule
ru$\rightarrow$en, fr & 321.9 & 196 & UN Corpus \\ \bottomrule
\end{tabular}
}
\caption{Statistics for seed parallel corpus. \textbf{en, zh, ru, fr, de, ja} is the abbreviations for English, Chinese, Russian, French, German and Japanese. \textbf{Avg. Tokens per Sample} refers to the average number of tokens in the source language input.}
\label{tab:fig_token}

\end{table}

\paragraph{Reasoning Strategies in Human Translation.} The six strategies are detailed as follows: 

(1) \textit{Hierarchical Translation:} Reflecting human translators' practice of segmenting complex texts into manageable components. By identifying essential elements first, the model translates each segment with greater contextual awareness, reducing errors from overlooked structural dependencies.

(2) \textit{Triangulation Translation:} Also known as relay translation~\citep{ringmar2012relay}, it employs an intermediate (pivot) language to bridge linguistic and cultural gaps~\citep{guyot2018translating}. When direct translation is difficult, pivoting enhances grammatical accuracy, vocabulary appropriateness, and cultural nuance.

(3) \textit{Back Translation:} Translating a target sentence back to the original language to detect and correct translation errors or inconsistencies~\citep{brislin2001back}, ensuring improved fluency and fidelity.

(4) \textit{Context-aware Translation:} Emphasizing the broader context (e.g., paragraph or document) in translation decisions~\citep{cabezas2023use,almanna2015contextualizing}, this strategy effectively preserves idiomatic expressions, cultural nuances, and domain-specific terminologies.

(5) \textit{Translation Explanation:} Providing rationales for translation choices when multiple interpretations are possible~\citep{buhler2002translation}, enhancing transparency and allowing for more informed and justified translation outcomes.

(6) \textit{Structural Transformation:} Systematically adjusting sentence structures to adhere to target language syntactic norms, essential for translations between structurally distinct languages such as English to Japanese or German. This ensures grammatical correctness and syntactic appropriateness.

\paragraph{Construction of CoT Templates from Human Experience.}

Drawing upon the six core translation reasoning strategies, we abstracted structured CoT templates directly from human translator's workflows, formulating each strategy into detailed, step-by-step instructions to enable LLMs to consistently instantiate explicit reasoning trajectories.

For example, the \textit{Hierarchical Translation} template explicitly guides the LLM through several structured stages: initially analyzing the source sentence to identify core linguistic elements (subject, verb, object), then performing a preliminary translation specifically focused on these core elements to ensure foundational accuracy. Subsequently, the model reviews the translation to identify areas needing refinement, such as word choice, grammatical correctness, or contextual coherence, and iteratively improves the translation's fluency and readability. Finally, it validates that the refined translation accurately preserves the original meaning and aligns well with the broader textual context. 
Similar detailed instructions were systematically developed for each of the other five templates, as fully documented in Appendix~\ref{sec:appendix B}. Each template has been meticulously crafted and calibrated by language experts to authentically reflect human translation workflows, enhancing transparency, and facilitating clear and effective instantiation by LLMs.

\paragraph{Incorporating CoTs into Parallel Corpus}
With the calibrated templates, we use advanced LLMs to generate instantiated translation CoTs for each sample within our seed dataset, leading to a total of 2k parallel pairs with translation CoTs. Each sample contains a source sentence and a target sentence, along with a reasoning trajectory revealing how the target translation reference is obtained step-by-step from the source sentence. Manual inspection by language experts are conducted to control the quality of generated training samples. 
Specifically, during the instantiation process, we utilize the Role-Play Prompting technique, as described by \citep{kong2023better}, to enable the LLM to simulate human-like reflection in translating the source sentence. The prompt guides the model through the instantiation process, strictly following the reasoning steps outlined in the CoT template. The detailed prompt is provided in Appendix~\ref{sec:appendix C}.

\subsection{Translation Reasoning Learning}  \label{sec: Translation Reasoning Learning}

\paragraph{Stage I: SFT with Reasoning Dataset as Cold Start}

In the process of \textit{translation reasoning learning}, we initiate the training process with a cold start phase, leveraging the MT Reasoning Dataset in Section~\ref{sec:build cot dataset} to fine-tune the model. This approach draws inspiration from Deepseek-R1 training strategies \citep{deepseek2025r1}. The cold start phase establishes a foundational layer of reasoning capabilities, which are further refined in subsequent phases of learning.  More importantly, human professional knowledge of translation can be injected into the model via step-by-step examples in the cold start phase, facilitating convergence of subsequent learning.

\paragraph{Stage II: Self-Evolving with RL}

This part introduces a self-evolving RL mechanism that allows the model to progressively refine its reasoning skills over time after the cold start phase. Notably, only the 2k training parallel corpus in Table~\ref{tab:fig_token} without CoT are used in this phase. This iterative process facilitates the continuous improvement of translation strategies, ensuring the model learns both from immediate outputs (\emph{i.e.}, token-level predictions) and adapts based on long-term feedback (\emph{i.e.}, overall translation quality). The reward calculation and learning strategy for RL are detailed below:

\textit{RL Reward Modeling:} For the RL stage, we have designed a reward modeling system consisting of two distinct types of rewards: 
\textit{(a) Format Reward:} This reward encourages the model to adhere to a structured response format. We use regular-expression extraction to ensure that the reasoning process is placed within the `<think></think>` tags and the final translation is enclosed within the `<answer></answer>` tags.
\textit{(b) Answer Reward:} The answer reward evaluates the quality of the translation in the model output. To objectively assess translation quality, we employ COMET~\citep{rei-etal-2020-comet}, an evaluation metric that compares the translation of the model with a reference.

However, several challenges arise during the design of this system: \textbf{a}. In practical applications, relatively large negative rewards cause the agent to become overly conservative, potentially trapping it in local optima ~\citep{schulman2017proximal}. \textbf{b}. When the scale of translation reward is much lower than the format reward, the model prioritizes format correctness over translation quality, leading to suboptimal behavior. 
To address these challenges, we first set the lower bound of the reward functions to 0 to prevent negative rewards. To further balance the rewards, we set the upper limit of the format reward to 0.2, as suggested by \citep{kocmi2024preliminary} for cases of low translation quality. This ensures that the format reward does not overly interfere with the translation quality, allowing the model to focus on both translation quality and formatting. Thus, the reward system \( r \) is computed as:

\begin{equation}\label{eq1}
r\_form = 
\begin{cases}
0.2 & \text{if the format is correct} \\
0 & \text{if the format is incorrect}
\end{cases}
\end{equation}

\begin{equation}\label{eq2}
r\_ans = 
\begin{cases}
0 & \text{if } x \leq 0 \\
Com(trans, ref) & \text{if } x > 0
\end{cases}
\end{equation}

\begin{equation}\label{eq3}
r = 
\begin{cases}
0 & \text{if } r\_form = 0 \\
r\_ans + r\_form & \text{if } r\_form > 0
\end{cases}
\end{equation}

Here, \( x\) donates Com(trans, ref), and Com donates the COMET-20 score, trans is the generated translation, ref is the reference translation. \( r_{\text{form}} \) and \( r_{\text{ans}} \) are Format Reward and Answer Reward.

\paragraph{RL Strategy}  

For the RL phase, we employ the Group Relative Policy Optimization (GRPO) algorithm~\citep{shao2024deepseekmath}, which was first publicly proposed to optimize reasoning LLMs using complex rewards without a critic model, employing group-based advantage estimation.

\section{Experiment}
\label{sec:exp}

\begin{table*}[t]
    \centering
    \resizebox{\linewidth}{!}{
        \begin{tabular}{l@{\hskip 0.05in}c@{\hskip 0.05in}c@{\hskip 0.05in}c@{\hskip 0.05in}c@{\hskip 0.05in}c@{\hskip 0.05in}c@{\hskip 0.05in}c@{\hskip 0.05in}c@{\hskip 0.05in}c@{\hskip 0.05in}c@{\hskip 0.05in}c@{\hskip 0.05in}c@{\hskip 0.05in}c@{\hskip 0.05in}c@{\hskip 0.05in}c@{\hskip 0.05in}c@{\hskip 0.05in}c@{\hskip 0.05in}c@{\hskip 0.05in}c@{\hskip 0.05in}c@{\hskip 0.05in}c}
            \toprule
    \multirow{2}{*}{\textbf{Models}} & \multicolumn{5}{c}{\hspace{-1.5em}\textbf{xx2en}}       & \multicolumn{5}{c}{\hspace{-0.2em}\textbf{en2xx}} & \multicolumn{5}{c}{\hspace{-0.2em}\textbf{xx2zh}} & \multicolumn{5}{c}{\hspace{-0.2em}\textbf{zh2xx}} & \textbf{avg}\\
    \cmidrule(l{0.3em}r{1em}){2-6} \cmidrule(l{0.3em}r{1em}){7-11} \cmidrule(l{0.3em}r{1em}){12-16} \cmidrule(l{0.3em}r{1em}){17-21} 
        & \textbf{zh} & \textbf{ja} & \textbf{ru} & \textbf{fr} & \textbf{de} & \textbf{zh} & \textbf{ja} & \textbf{ru} & \textbf{fr} & \textbf{de} & \textbf{en} & \textbf{ja} & \textbf{ru} & \textbf{fr} & \textbf{de} & \textbf{en} & \textbf{ja} & \textbf{ru} & \textbf{fr} & \textbf{de} & \textbf{-} \\ \midrule
    \multicolumn{21}{c}{\textbf{General-purpose LLMs}} \\
    \midrule
    Q2.5$^{\mathrm{a}}$-7b & 0.700 & 0.659 & 0.649 & \textbf{0.800} & 0.747 & 0.624 & 0.435 & 0.538 & 0.759 & 0.559 &  0.624 & 0.583 & 0.525 & 0.565 & 0.527 & 0.699 & 0.531 & 0.489 & 0.555 & 0.475 & 0.602 \\
    Q2.5-7b-Ins$^{\mathrm{b}}$ & 0.673 & 0.641 & 0.556 & 0.736 & 0.604 & 0.622 & 0.452 & 0.343 & 0.495 & 0.453 & 0.622 & 0.550 & 0.465 & 0.531 & 0.473 & 0.672 & 0.450 & 0.087 & 0.152 & 0.280 & 0.493 \\
    Q2.5$^{\mathrm{a}}$-14b& 0.649 & 0.600 & 0.598 & 0.751 & 0.692 & 0.408 & -0.059 & 0.399 & 0.577 & 0.458 & 0.408 & 0.505 & 0.425 & 0.473 & 0.450 & 0.649 & 0.004 & 0.388 & 0.433 & 0.355 & 0.458 \\
    la3.1$^{\mathrm{c}}$-8b-Ins$^{\mathrm{b}}$ & 0.572 & 0.640 & 0.412 & 0.379 & 0.388 & 0.332 & 0.592 & 0.621 & 0.742 & 0.559 & 0.332 & 0.421 & 0.318 & 0.056 & 0.181 & 0.572 & 0.581 & 0.562 & 0.544 & 0.461 & 0.463 \\

    \midrule
    \multicolumn{21}{c}{\textbf{Reasoning LLMs}} \\
    \hline
    DS-R1-D$^{\mathrm{d}}$ & 0.572 & 0.377 & 0.441 & 0.684 & 0.577 & 0.452 & -1.020 & -0.851 & 0.033 & -0.454 & 0.452 & 0.009 & 0.199 & 0.290 & 0.206 & 0.571 & -0.621 & -0.903 & -0.222 & -0.580 & 0.011 \\
    DS-R1-DM$^{\mathrm{e}}$ & 0.572 & 0.321 & 0.437 & 0.689 & 0.577 & 0.504 & -0.662 & -0.508 & 0.077 & -0.299 & 0.504 & -0.014 & 0.266 & 0.292 & 0.193 & 0.572 & -0.474 & -0.744 & -0.279 & -0.538 & 0.074 \\
    Marco-o1 & 0.687 & 0.627 & 0.639 & 0.784 & 0.731
         & 0.634 & 0.539 & 0.595 & 0.746 & 0.570
         & 0.634 & 0.588 & 0.522 & 0.577 & 0.557
         & 0.687 & 0.556 & 0.518 & 0.537 & 0.467
         & 0.610 \\
    \midrule
    \multicolumn{21}{c}{\textbf{Translation Models}} \\
    \hline
    ParroT-7b & 0.476 & 0.322 & 0.559 & 0.740 & 0.688 & 0.176 & -0.600 & -0.455 & 0.480 & 0.476 &
      0.176 & -0.602 & -0.484 & -0.525 & -0.554 & 0.476 & -0.893 & -1.577 & -0.437 & -0.880 & -0.122 \\
    NLLB
      & 0.439 & 0.389 & 0.180 & 0.171 & 0.312
      & -0.047 & -0.052 & 0.413 & 0.443 & 0.265
      & -0.047 & 0.051 & -0.242 & -0.321 & -0.169
      & 0.439 & 0.075 & 0.266 & 0.186 & 0.083
      & 0.142 \\
    \midrule
    
    \textbf{R1-T1} & \textbf{0.707} & \textbf{0.659} & \textbf{0.650} & 0.797 & \textbf{0.748} & \textbf{0.672} & \textbf{0.644} & \textbf{0.642} & \textbf{0.771} & \textbf{0.597} & \textbf{0.672} & \textbf{0.624} & \textbf{0.555} & \textbf{0.608} & \textbf{0.596} & \textbf{0.707} & \textbf{0.645} & \textbf{0.601} & \textbf{0.599} & \textbf{0.529} & \textbf{0.651} \\
    \bottomrule 
    \end{tabular}
    }
    \caption{R1-T1's performance on \textbf{Trained} languages in Flores-101. \textbf{xx2en} (or \textbf{xx2zh}) means the translation direction is from language \textbf{xx} to English (or Chinese). \textbf{Q2.5} means Qwen2.5. \textbf{Ins} means Instruction. \textbf{DS-R1-D(M)} means DeepSeek-R1-Distill-Qwen-7B(-Multilingual). zh,ja,ru,fr,de refer to Chinese, Japanese, Russian, French, German. Other definition of notations are the same as Table \ref{tab:unseen_flores}}
    \label{tab:trained_flores}
\end{table*}

In this section, we provide an overall evaluation for R1-T1, covering general MT, domain-specific MT, ablation study, and the case study of self-evolution translation CoT during RL.

\subsection{Experiment Setting} \label{sec::setting}

\paragraph{Implement Details} 

we selected Qwen2.5-7B-Instruct as a backbone model due to its superior multilingual performance among open source models with fewer than 10 billion parameters~\citep{Yang2024Qwen25TR}. This choice helps minimize potential negative impacts from insufficient base model capabilities. The dataset was randomly split into a training set and a validation set at a ratio of 9:1. For GRPO implementation, we develop it based on VeRL Platform\footnote{\url{https://github.com/volcengine/verl}}. The training configuration consists of 3 epochs, a learning rate of $3 \times 10^{-7}$, a batch size of 8, and 16 rollouts. For SFT with CoTs, we used 2 epochs of full parameter fine-tuning with a learning rate of $1 \times 10^{-4}$.

\paragraph{Evaluation Dataset} 

We evaluated our model under general purpose and domain-specific benchmarks. For general multilingual translation, we employ \textit{Flores-101}~\citep{Goyal2021TheFE}, which offers extensive coverage of diverse languages across a wide range of topics, making it suitable for evaluating general MT capabilities. In addition, we added four domain-specific benchmarks to evaluate robustness under challenging translation scenarios. 
\textit{CommonsenseMT}~\citep{he2025box} evaluates models’ ability to resolve lexical and syntactic ambiguities using commonsense knowledge in Chinese-English translation. 
\textit{RTT}~\citep{zhang2023understanding} assesses MT robustness to terminology constraints using an English-German benchmark with multi-term inputs that challenge constraint fidelity.
\textit{CultureMT}~\citep{yao2023benchmarking} measures cultural translation adequacy via a multilingual benchmark annotated with culture-specific items requiring pragmatic adaptation.
\textit{DRT Literature}~\citep{jiaan2024drt} offers literary texts rich in figurative language, testing models’ ability to preserve metaphors, similes, and stylistic fidelity. To avoid data leakage, Table ~\ref{tab:token_similarity} summarizes the token-level overlap between these evaluation sets and our training data, which produced a similarity score of less than 0.15.

\begin{table}[htbp]
\centering
\resizebox{1.0\linewidth}{!} {%
\begin{tabular}{lcccccc}
\toprule
\textbf{Dataset} & \textbf{ru} & \textbf{fr} & \textbf{de} & \textbf{ja} & \textbf{zh} & \textbf{en}\\
\midrule
Flores101  & 0.142 & 0.137 & 0.097 & 0.113 & 0.110 & 0.097 \\
Common sense  & - & - & - & - & 0.071 & 0.089 \\
Terminology  & - & - & 0.106 & - & - & 0.108 \\
Culture  & - & 0.135 & 0.106 & - & 0.102 & 0.123 \\
Literature  & - & - & - & - & 0.102 & 0.105 \\

\bottomrule
\end{tabular}
}
\caption{Similarity between test set and our training set.}
\label{tab:token_similarity}
\end{table}

\paragraph{Metric} We use both automatic and manual evaluation. For automatic evaluation, we employ COMET-20~\citep{rei-etal-2020-comet} to measure the quality of translation hypotheses given references. It has shown a strong correlation with human judgments in shared tasks like WMT20 and WMT23~\citep{freitag2022results, freitag2023results}, and is robust in diverse languages and domains~\citep{zouhar2024pitfalls}.
For human evaluation, we cooperated with a group of professional human translators from the language service center of a top-tier cooperation. And the evaluation was conducted strictly under their rigorous metric, which is similar to DA \citep{barrault2019findings} and MQM \citep{freitag2021experts}. Specifically, the annotators first rate each translation output on two dimensions: \textit{accuracy}, \textit{fluency}, using a 5-point Likert scale (1: poor, 5: excellent) and finally give a DA score. Besides that, the evaluation criteria and the error taxonomy guiding annotators are adapted from the MQM framework, incorporating fine-grained error types such as mistranslation, omission, grammar, and word order.

\subsection{Main Results}   
Our main experiment focuses on model performance of multilingual MT, investigating how well its performance in both trained and unseen languages against existing approaches.

\begin{table*}[t]

    \centering
    \resizebox{\linewidth}{!}{
        \begin{tabular}{l@{\hskip 0.05in}c@{\hskip 0.05in}c@{\hskip 0.05in}c@{\hskip 0.05in}c@{\hskip 0.05in}c@{\hskip 0.05in}c@{\hskip 0.05in}c@{\hskip 0.05in}c@{\hskip 0.05in}c@{\hskip 0.05in}c@{\hskip 0.05in}c@{\hskip 0.05in}c@{\hskip 0.05in}c@{\hskip 0.05in}c@{\hskip 0.05in}c@{\hskip 0.05in}c@{\hskip 0.05in}c@{\hskip 0.05in}c@{\hskip 0.05in}c@{\hskip 0.05in}c@{\hskip 0.05in}c}
            \toprule
    \multirow{2}{*}{\textbf{Models}} & \multicolumn{5}{c}{\hspace{-1.5em}\textbf{xx2en}}       & \multicolumn{5}{c}{\hspace{-0.2em}\textbf{en2xx}} & \multicolumn{5}{c}{\hspace{-0.2em}\textbf{xx2zh}} & \multicolumn{5}{c}{\hspace{-0.2em}\textbf{zh2xx}} & \textbf{avg}\\
    \cmidrule(l{0.3em}r{1em}){2-6} \cmidrule(l{0.3em}r{1em}){7-11} \cmidrule(l{0.3em}r{1em}){12-16} \cmidrule(l{0.3em}r{1em}){17-21} 
        & \textbf{th} & \textbf{nl} & \textbf{vi} & \textbf{tr} & \textbf{cs} & \textbf{th} & \textbf{nl} & \textbf{vi} & \textbf{tr} & \textbf{cs} & \textbf{th} & \textbf{nl} & \textbf{vi} & \textbf{tr} & \textbf{cs} & \textbf{th} & \textbf{nl} & \textbf{vi} & \textbf{tr} & \textbf{cs} & \textbf{-} \\ \midrule
    \multicolumn{21}{c}{\textbf{General-purpose LLMs}} \\
    \midrule

    Q2.5$^{\mathrm{a}}$-7b & \textbf{0.667} & 0.482 & 0.715 & 0.441 & 0.701 & 0.438 & \textbf{0.672} & 0.619 & 0.381 & 0.513 & 0.515 & 0.501 & 0.578 & 0.368 & 0.521 & 0.355 & 0.409 & \textbf{0.589} & 0.256 & 0.410 & 0.507 \\
    Q2.5-7b-Ins$^{\mathrm{b}}$ & 0.566 & 0.580 & 0.536 & 0.391 & 0.601 & 0.148 & 0.026 & 0.226 & 0.185 & 0.333 & 0.434 & 0.418 & 0.456 & 0.301 & 0.470 & 0.012 & -0.046 & -0.065 & -0.237 & 0.075 & 0.270 \\
    Q2.5$^{\mathrm{a}}$-14b & 0.601 & 0.599 & 0.616 & 0.563 & 0.624 & -0.173 & 0.220 & 0.354 & -0.918 & -0.075 & 0.505 & 0.358 & 0.466 & 0.362 & 0.141 & -0.076 & 0.134 & 0.339 & -1.041 & -0.014 & 0.184 \\
    la3.1$^{\mathrm{c}}$-8b-Ins$^{\mathrm{b}}$ & 0.480 & 0.299 & 0.531 & 0.266 & 0.534 & 0.419 & 0.478 & 0.609 & 0.452 & 0.472 & 0.033 & 0.227 & 0.321 & 0.258 & 0.337 & 0.329 & 0.372 & 0.549 & 0.220 & 0.351 & 0.377 \\
    \midrule
    \multicolumn{21}{c}{\textbf{Reasoning LLMs}} \\
    \midrule
    DS-R1-D$^{\mathrm{c}}$ & 0.143 & 0.408 & 0.217 & 0.156 & 0.323 & -1.222 & -0.777 & -0.925 & -0.812 & -0.900 & -0.194 & -0.016 & -0.160 & -0.295 & -0.049 & -0.986 & -0.786 & -0.805 & -0.915 & -0.979 & -0.429 \\
    DS-R1-DM$^{\mathrm{d}}$ & 0.161 & 0.415 & 0.181 & 0.023 & 0.331 & -0.930 & -0.593 & -0.538 & -0.715 & -0.767 & -0.478 & -0.072 & -0.467 & -0.445 & 0.063 & -0.868 & -0.739 & -0.691 & -0.901 & -0.840 & -0.394 \\
    Marco-o1 & 0.644 & 0.666 & 0.699 & 0.545 & 0.699
         & 0.389 & 0.484 & 0.607 & 0.459 & 0.502
         & 0.535 & 0.514 & 0.570 & 0.418 & 0.533
         & 0.296 & 0.393 & 0.534 & 0.191 & 0.443
         & 0.506  \\
    \midrule
    \multicolumn{21}{c}{\textbf{Translation Models}} \\
    \midrule
    ParroT-7b &-0.873 &  0.626 & -0.546 & -0.183 &  0.622 & -1.069 &  0.326 & -0.976 & -1.067 &  0.257 &
             -1.364 & -0.585 & -1.184 & -0.874 & -0.536 & -0.878 & -0.580 & -0.560 & -1.080 & -0.954 & -0.574 \\
    NLLB   & 0.170 & 0.244 & 0.373 & 0.400 & 0.182 & -0.506 & 0.287 & 0.336 & 0.406 & 0.473 
           & -0.185 & -0.281 & -0.015 & -0.086 & -0.312 & -0.356 & 0.100 & 0.274 & 0.127 & 0.205 & 0.092 \\ 
    \midrule
    \textbf{R1-T1} & 0.659 & \textbf{0.687} & \textbf{0.720} & \textbf{0.677} & \textbf{0.713} & \textbf{0.468} & 0.521 & \textbf{0.644} & \textbf{0.552} & \textbf{0.551} & \textbf{0.573} & \textbf{0.548} & \textbf{0.610} & \textbf{0.530} & \textbf{0.571} & \textbf{0.411} & \textbf{0.458} & 0.5593 & \textbf{0.387} & \textbf{0.491} & \textbf{0.567} \\
    \bottomrule 
    \end{tabular}
    }
    \caption{R1-T1's performance on \textbf{Unseen} languages in Flores-101. \textbf{th}, \textbf{nl}, \textbf{vi}, \textbf{tr}, and \textbf{cs} refer to Thai, Dutch, Vietnamese, Turkish and Czech. Defination of notations are the same as Table \ref{tab:trained_flores}}
    \label{tab:unseen_flores}
\end{table*}

\paragraph{Baselines} 
We compare our R1-T1 against several famous models holding multilingual MT capability, including (1) General-Purpose LLMs: \textit{Qwen2.5-7b-base}, \textit{Qwen2.5-7B-Instruct}, \textit{Qwen2.5-14B}, \textit{Llama3.1-8B}; (2) Reasoning-LLMs: \textit{DeepSeek-R1-Distill-Qwen-7B}, \textit{DeepSeek-R1-Distill-Qwen-7B-Multilingual}~\footnote{\url{https://huggingface.co/lightblue/DeepSeek-R1-Distill-Qwen-7B-Multilingual}}, Marco-o1 ~\citep{zhao2024marco}; (3) Translation models: \textit{ParroT-7b} ~\citep{jiao2023parrot} and \textit{NLLB} ~\citep{costa2022no}.

\paragraph{Result of Trained Languages}

As shown in Table \ref{tab:trained_flores}, R1-T1 achieves the highest average COMET score (0.651) in all supervised translation directions in Flores-101, demonstrating robust performance in six high-resource languages. Compared to general-purpose LLMs like Qwen2.5 series and LLaMA3.1-8B-Instruct, R1-T1 consistently achieves higher COMET scores, particularly in challenging language pairs such as ru2en and ja2en. Reasoning-oriented models exhibit inconsistent or unstable performance, with negative or near-zero scores in certain directions, suggesting their reasoning mechanisms alone are insufficient for consistently strong multilingual translation. Translation-specific models perform notably lower, highlighting the limitations of task-specific training without reasoning augmentation. The strong and balanced performance of R1-T1 validates the effectiveness of our reasoning-enhanced training paradigm.

\paragraph{Result of Unseen Languages}

Table~\ref{tab:unseen_flores} shows that R1-T1 substantially outperforms all baseline models on unseen language translation, highlighting its robust generalization. General-purpose models (e.g., Qwen2.5-7B-base) perform competitively in easier directions like Thai$\to$English but struggle with linguistically distant languages such as Vietnamese or Czech. Reasoning-oriented and translation-specific baselines display unstable and often negative scores, reflecting their limited zero-shot transfer capability, especially on unseen languages. In contrast, R1-T1, benefiting from RL-enhanced reasoning, consistently achieves superior COMET scores, notably improving Thai$\to$English translation (0.659 vs.~0.566). This confirms that our reasoning-enhanced training effectively boosts translation quality for unseen languages.

\subsection{Evaluation on Domain-specific MT} %

As stated in Section~\ref{sec::setting}, we evaluate R1-T1 across four challenging domains: \emph{commonsense}, \emph{terminology}, \emph{culture}, and \emph{literature}. The results, shown in Table~\ref{tab:domain_MT}, indicate that R1-T1 consistently outperforms all baseline models in average translation quality across these domains. The only exception is \textit{DRT-o1-7b}, which achieves slightly higher scores in the literature for en2zh, possibly due to its extensive exposure to en2zh literary data during training. These findings align with our central hypothesis that RL-augmented reasoning training enables better generalization to challenging domain tasks. For common sense or terminology scenarios, reasoning encourages the model to better preserve critical expressions and maintain contextual accuracy. In cultural or literary, CoT guidance helps the model disambiguate nuanced expressions and interpret figurative symbols within the appropriate situational context.

\begin{table}[t]
\centering
\renewcommand{\arraystretch}{1.05}
\setlength{\tabcolsep}{4pt}
\fontsize{9pt}{10pt}\selectfont

\resizebox{0.7\columnwidth}{!}{
\begin{tabular}{lcccc}
\toprule
\multirow{2}{*}{\textbf{Models}} & 
\multicolumn{1}{c}{\textbf{Cmn.}} & 
\multicolumn{2}{c}{\textbf{Literature}} & 
\multicolumn{1}{c}{\textbf{Term}} \\
\cmidrule(lr){2-2} \cmidrule(lr){3-4} \cmidrule(lr){5-5}
& \textbf{en2zh} & \textbf{en2zh} & \textbf{zh2en} & \textbf{en2de} \\
\midrule
Q\textsuperscript{a}2.5-7b-Ins\textsuperscript{b} & 0.481 & 0.187 & 0.070 & 0.402 \\
Q\textsuperscript{a}2.5-14b & 0.384 & -0.284 & 0.140 & 0.362 \\
la3.1\textsuperscript{c}-8b-Ins\textsuperscript{b} & 0.337 & -0.162 & 0.106 & \textbf{0.545} \\
Marco-o1 & 0.529 & 0.238 & 0.195 & 0.457 \\
DRT-o1-7b & -- & \textbf{0.383} & -0.866 & -- \\
\midrule
\textbf{R1-T1} & \textbf{0.584} & 0.313 & \textbf{0.223} & 0.506 \\
\bottomrule
\end{tabular}
}

\vspace{2pt} 
\resizebox{\columnwidth}{!}{
\begin{tabular}{lccccccc}
\toprule
\multirow{2}{*}{\textbf{Models}} & 
\multicolumn{6}{c}{\textbf{Culture}} & 
\multirow{2}{*}{\textbf{avg}} \\
\cmidrule(lr){2-7}
& \textbf{zh2en} & \textbf{en2zh} & \textbf{es2en} & \textbf{en2es} & \textbf{fr2en} & \textbf{en2fr} & \\
\midrule
Q\textsuperscript{a}2.5-7b-Ins\textsuperscript{b} & 0.342 & 0.401 & 0.576 & 0.523 & 0.235 & 0.197 & 0.341 \\
Q\textsuperscript{a}2.5-14b & 0.317 & 0.018 & 0.494 & 0.353 & 0.196 & 0.036 & 0.202 \\
la3.1\textsuperscript{c}-8b-Ins\textsuperscript{b} & 0.325 & 0.305 & 0.375 & 0.424 & 0.122 & 0.174 & 0.255 \\
Marco-o1 & 0.351 & 0.394 & 0.545 & 0.523 & 0.215 & 0.201 & 0.365 \\
DRT-o1-7b & -- & -- & -- & -- & -- & -- & -- \\
\midrule
\textbf{R1-T1} & \textbf{0.402} & \textbf{0.455} & \textbf{0.580} & \textbf{0.545} & \textbf{0.246} & \textbf{0.230} & \textbf{0.501} \\
\bottomrule
\end{tabular}
}

\caption{R1-T1’s performance on Domain-specific MT. Cmn. means Common sense.}
\label{tab:domain_MT}
\end{table}

\subsection{Ablation Study} 

\paragraph{Ablation on the RL incentivization for MT} \label{sec:ablation}

We trained and compared models with identical training settings except for distinct training paradigms and evaluated their translation performance on various benchmarks. Specifically, we consider five settings:
(1) \textit{Q2.5-7b}: Original Qwen2.5-7b-instruct without additional fine-tuning.
(2) \textit{Q2.5-7b + SFT}: SFT applied directly using the 2k parallel translation corpora without CoT.
(3) \textit{Q2.5-7b + RL}: Directly conducting stage II of the \textit{translation reasoning learning} (i.e., RL) on Qwen2.5-7b-instruct.
(4) \textit{Q2.5-7b (w/ human CoT) + SFT}: Conducting only stage I (i.e., SFT with Human CoT). 
(5) \textit{Q2.5-7b (w/ human CoT) + SFT + RL}: The full R1-T1 training pipeline with both stages.

As demonstrated in Table~\ref{tab:ablation_study}, the following key findings emerge clearly:
a. Group (3) (puerly RL) significantly improves translation quality compared to both baseline and pure SFT methods, highlighting that the self-evolving RL enhances general MT performance.
b. Directly applying Human CoT data through SFT, (i.e., group (4)), leads to notably inferior results across benchmarks. This indicates that strictly enforcing a fixed reasoning paradigm without incentive flexibility degrades the model’s generalization.
c. When employing the Human CoT-supervised variant as a cold-start initialization for RL training (group (5)), the resulting model achieves the best overall performance. This shows that combining structured Human CoT guidance with RL-based exploratory training effectively equips the model with robust reasoning priors and adaptive reasoning capabilities.

\paragraph{Ablation on human expert-aligned CoT data}
We further examine the contribution of proposed MT Reasoning Dataset (with human CoT) by comparing (5) with (6) \textit{Q2.5-7b (w/o human CoT) + SFT + RL}:
Training with the same setting as R1-T1, but replacing the human CoT with a purely LLM-generated CoT dataset for cold start. Results in Table~\ref{tab:ablation_study} reveal substantial improvements in translation quality when incorporating Human CoT as cold start training, indicating that human prior knowledge significantly strengthens the model’s reasoning capability and robustness across diverse translation tasks.

\begin{table}[t]
\centering
\resizebox{\linewidth}{!}{
\begin{tabular}{l@{\hskip 0.05in}c@{\hskip 0.05in}c@{\hskip 0.05in}c@{\hskip 0.05in}c@{\hskip 0.05in}c@{\hskip 0.05in}c}
\toprule
\textbf{Models} & \textbf{F101} & \textbf{Cmn.} & \textbf{Lit} & \textbf{Term} & \textbf{Culture} \\
\\
\midrule
Q2.5-7b                               & 0.554 & 0.481 & 0.129   & 0.402 & 0.379 \\[0.1em]
Q2.5-7b + SFT                         & 0.566 & 0.528 & 0.190 & 0.481 & 0.407 \\[0.1em]
Q2.5-7b + RL                          & 0.574 & 0.582 & 0.267 & 0.473 & 0.398 \\[0.1em]
Q2.5-7b (w/ human CoT) + SFT          & 0.435 & 0.525 & 0.454 & 0.480 & 0.403 \\[0.1em]
Q2.5-7b (w/o human CoT) + SFT + RL    & 0.528 & --    & --    & --    & --    \\[0.1em]
\hdashline
\textbf{R1-T1} &&&&&\\
Q2.5-7b (w/ human CoT) + SFT + RL     & \textbf{0.609} & \textbf{0.584} & \textbf{0.269} & \textbf{0.506} & \textbf{0.410} \\[0.1em]
\bottomrule
\end{tabular}}
\caption{Ablation study.  F101 = Flores‑101; CmnS = Common Sense; Lit = Literature}
\label{tab:ablation_study}
\end{table}

\subsection{Human Evaluation} 
As automatic metrics may not capture nuanced translation defects~\citep{freitag2021experts,agrawal2024can}, human evaluations by professional translators are essential. Following the rigorous evaluation criteria outlined in Section~\ref{sec::setting}, we conducted a comprehensive human evaluation for R1-T1. Specifically, we uniformly sampled 720 translation instances from the five test sets introduced previously.  
Table~\ref{tab:human_eval} summarizes the human evaluation results. R1-T1 consistently achieves superior DA and Fluency scores in all evaluated domains. 
Notable improvements occur in domains requiring precise terminology usage (\textit{Term}) and nuanced figurative comprehension (\textit{Literature}). These domains highlight subtle differences, such as fluency in idiomatic expression and contextual precision, that are typically perceivable only through human evaluation, underscoring the robustness and linguistic sophistication of R1-T1. Further translator comments are presented in the Appendix ~\ref{sec:appendix E}.

\begin{table}[t]
  \centering
  \renewcommand{\arraystretch}{1.01}
  \setlength{\tabcolsep}{1pt}
\resizebox{0.9\linewidth}{!}{

  \begin{tabular}{@{}l@{\hskip 0.03in}*{9}{c@{\hskip 0.04in}}}
    \toprule
    \multirow{2}{*}{\textbf{Models}} &
      \multicolumn{3}{c}{\textbf{Flores-101}} &
      \multicolumn{3}{c}{\textbf{Common sense}} &
      \multicolumn{3}{c}{\textbf{Literature}} \\
    \cmidrule(l{0.2em}r{0.4em}){2-4}
    \cmidrule(l{0.2em}r{0.4em}){5-7}
    \cmidrule(l{0.2em}r{0.4em}){8-10}
    & Flu & Acc & DA 
    & Flu & Acc & DA 
    & Flu & Acc & DA \\
    \midrule
    Q2.5-7B-SFT & 
    4.41 & 4.25 & 86.81 &  
    4.45 & \textbf{4.85} & 90.75 &
    76.80 & 3.95 & 4.45 \\
    \textbf{R1-T1} &
    \textbf{4.41} & \textbf{4.38} & \textbf{88.36} &
    \textbf{4.80} & 4.65 & \textbf{93.50} &
    \textbf{76.45} & \textbf{4.00} & \textbf{4.55} \\
    \bottomrule
  \end{tabular}}

  \vspace{4pt}
  \resizebox{0.68\linewidth}{!}{
  \begin{tabular}{@{}l@{\hskip 0.03in}*{6}{c@{\hskip 0.04in}}}
    \toprule
    \multirow{2}{*}{\textbf{Models}} &
      \multicolumn{3}{c}{\textbf{Term}} &
      \multicolumn{3}{c}{\textbf{Culture}} \\
    \cmidrule(l{0.2em}r{0.4em}){2-4}
    \cmidrule(l{0.2em}r{0.4em}){5-7}
    & Flu & Acc & DA 
    & Flu & Acc & DA \\
    \midrule
    Q2.5-7B-SFT & 
    4.50 & 4.00 & 88.00 &
    4.67 & 4.13 & 84.10 \\
    \textbf{R1-T1} &
    \textbf{4.80} & \textbf{4.70} & \textbf{93.10} & 
    \textbf{4.72} & \textbf{4.17} & \textbf{84.38} \\
    \bottomrule
  \end{tabular}}

  \caption{Human evaluation results across five domains. Fluency (Flu.), Accuracy (Acc.), and Direct Assessment (DA). Higher is better.}
  \label{tab:human_eval}
\end{table}

\subsection{CoT Self-evolution Analysis}

To provide a more intuitive evaluation of the CoT self-evolution mechanism, we conduct a visualization for the translation of a complex sentence from Chinese to English, with and without the integration of self-evolving CoTs in Fig. \ref{fig1}.  In both instances, the translated sentence correctly conveys the meaning, but the version utilizing self-evolution provides more nuanced translations. The description without CoTs includes generic phrasing, such as "the meeting was held," whereas the self-evolved CoT version includes additional context, such as "the critical meeting was held to address urgent concerns."

Through this analysis, it becomes evident that the self-evolving CoT improves the model's ability to adapt over time, refining its translation strategies. The self-evolving CoT allows the model to generate more diverse and accurate translations by incorporating expert-level reasoning strategies and incorporating feedback from prior outputs. This demonstrates that CoT self-evolution is effective in enhancing the model's reliability and its capacity to provide richer, context-aware translations across languages.

\section{Conclusion}
In this paper, we revealed the benefits of performing general MT tasks with inference-time reasoning, by combining human-aligned translation CoTs and RL strategies tailored for MT tasks. The success of R1-T1, our proposed MT framework, on MT tasks across multiple domains and languages reflects the necessity and wide existence of the reasoning processes in human translations, which are ignored by existing methods. Moreover, the self-evolved CoTs in translation reasoning learning provide R1-T1 with high flexibility in generalizing its translation ability to unseen languages, domains and tasks, as indicated by the improved translation performances in these unseen scenarios. Future work include incorporating more languages and CoTs in training and testing R1-T1 in more scenarios.  

\bibliography{R1-translator}

\begin{thebibliography}{56}
\providecommand{\natexlab}[1]{#1}

\bibitem[{Agrawal et~al.(2024)Agrawal, Farinhas, Rei, and Martins}]{agrawal2024can}
Sweta Agrawal, Ant{\'o}nio Farinhas, Ricardo Rei, and Andr{\'e}~FT Martins. 2024.
\newblock Can automatic metrics assess high-quality translations?
\newblock \emph{arXiv preprint arXiv:2405.18348}.

\bibitem[{Almanna(2015)}]{almanna2015contextualizing}
Ali Almanna. 2015.
\newblock \emph{Contextualizing translation theories: Aspects of Arabic--English interlingual communication}.
\newblock Cambridge Scholars Publishing.

\bibitem[{Ba{\~n}{\'o}n et~al.(2020)Ba{\~n}{\'o}n, Chen, Haddow, Heafield, Hoang, Espl{\`a}-Gomis, Forcada, Kamran, Kirefu, Koehn et~al.}]{banon2020paracrawl}
Marta Ba{\~n}{\'o}n, Pinzhen Chen, Barry Haddow, Kenneth Heafield, Hieu Hoang, Miquel Espl{\`a}-Gomis, Mikel Forcada, Amir Kamran, Faheem Kirefu, Philipp Koehn, et~al. 2020.
\newblock Paracrawl: Web-scale acquisition of parallel corpora.
\newblock Association for Computational Linguistics (ACL).

\bibitem[{Barrault et~al.(2019)Barrault, Bojar, Costa-Jussa, Federmann, Fishel, Graham, Haddow, Huck, Koehn, Malmasi et~al.}]{barrault2019findings}
Lo{\"\i}c Barrault, Ond{\v{r}}ej Bojar, Marta~R Costa-Jussa, Christian Federmann, Mark Fishel, Yvette Graham, Barry Haddow, Matthias Huck, Philipp Koehn, Shervin Malmasi, et~al. 2019.
\newblock Findings of the 2019 conference on machine translation (wmt19).
\newblock ACL.

\bibitem[{Bojar et~al.(2018)Bojar, Federmann, Fishel, Graham, Haddow, Huck, Koehn, and Monz}]{bojar-etal-2018-findings}
Ond{\v{r}}ej Bojar, Christian Federmann, Mark Fishel, Yvette Graham, Barry Haddow, Matthias Huck, Philipp Koehn, and Christof Monz. 2018.
\newblock \href {https://doi.org/10.18653/v1/W18-6401} {Findings of the 2018 conference on machine translation ({WMT}18)}.
\newblock In \emph{Proceedings of the Third Conference on Machine Translation: Shared Task Papers}, pages 272--303, Belgium, Brussels. Association for Computational Linguistics.

\bibitem[{Boughorbel et~al.(2024)Boughorbel, Parvez, and Hawasly}]{boughorbel2024improving}
Sabri Boughorbel, Md~Rizwan Parvez, and Majd Hawasly. 2024.
\newblock Improving language models trained on translated data with continual pre-training and dictionary learning analysis.
\newblock In \emph{Proceedings of The Second Arabic Natural Language Processing Conference}, pages 73--88.

\bibitem[{Brislin and Freimanis(2001)}]{brislin2001back}
Richard~W Brislin and Carolina Freimanis. 2001.
\newblock Back-translation.
\newblock \emph{An Encyclopaedia of Translation: Chinese-English, English-Chinese}, 22.

\bibitem[{B{\"u}hler(2002)}]{buhler2002translation}
Axel B{\"u}hler. 2002.
\newblock Translation as interpretation.
\newblock \emph{Translation studies: Perspectives on an emerging discipline}, pages 56--74.

\bibitem[{Cabezas-Garc{\'\i}a(2023)}]{cabezas2023use}
Melania Cabezas-Garc{\'\i}a. 2023.
\newblock The use of context in multiword-term translation.
\newblock \emph{Perspectives}, 31(2):365--382.

\bibitem[{Chu et~al.(2025{\natexlab{a}})Chu, Zhai, Yang, Tong, Xie, Schuurmans, Le, Levine, and Ma}]{chu2025sft}
Tianzhe Chu, Yuexiang Zhai, Jihan Yang, Shengbang Tong, Saining Xie, Dale Schuurmans, Quoc~V Le, Sergey Levine, and Yi~Ma. 2025{\natexlab{a}}.
\newblock Sft memorizes, rl generalizes: A comparative study of foundation model post-training.
\newblock \emph{arXiv preprint arXiv:2501.17161}.

\bibitem[{Chu et~al.(2025{\natexlab{b}})Chu, Tan, Xue, Wang, Mo, and Li}]{chu2025domaino1s}
Xu~Chu, Zhijie Tan, Hanlin Xue, Guanyu Wang, Tong Mo, and Weiping Li. 2025{\natexlab{b}}.
\newblock Domaino1s: Guiding llm reasoning for explainable answers in high-stakes domains.
\newblock \emph{arXiv preprint arXiv:2501.14431}.

\bibitem[{Costa-Juss{\`a} et~al.(2022)Costa-Juss{\`a}, Cross, {\c{C}}elebi, Elbayad, Heafield, Heffernan, Kalbassi, Lam, Licht, Maillard et~al.}]{costa2022no}
Marta~R Costa-Juss{\`a}, James Cross, Onur {\c{C}}elebi, Maha Elbayad, Kenneth Heafield, Kevin Heffernan, Elahe Kalbassi, Janice Lam, Daniel Licht, Jean Maillard, et~al. 2022.
\newblock No language left behind: Scaling human-centered machine translation.
\newblock \emph{arXiv preprint arXiv:2207.04672}.

\bibitem[{DeepSeek-AI(2025)}]{deepseek2025r1}
DeepSeek-AI. 2025.
\newblock Deepseek-r1: Incentivizing reasoning capability in llms via reinforcement learning.
\newblock In \emph{arXiv preprint arXiv:2501.12948}.

\bibitem[{Feng et~al.(2024)Feng, Zhang, Li, Liu, Lang, Feng, Wu, and Liu}]{feng2024improving}
Zhaopeng Feng, Yan Zhang, Hao Li, Wenqiang Liu, Jun Lang, Yang Feng, Jian Wu, and Zuozhu Liu. 2024.
\newblock Improving llm-based machine translation with systematic self-correction.
\newblock \emph{arXiv e-prints}, pages arXiv--2402.

\bibitem[{Freitag et~al.(2021)Freitag, Foster, Grangier, Ratnakar, Tan, and Macherey}]{freitag2021experts}
Markus Freitag, George Foster, David Grangier, Viresh Ratnakar, Qijun Tan, and Wolfgang Macherey. 2021.
\newblock Experts, errors, and context: A large-scale study of human evaluation for machine translation.
\newblock \emph{Transactions of the Association for Computational Linguistics}, 9:1460--1474.

\bibitem[{Freitag et~al.(2023)Freitag, Mathur, Lo, Avramidis, Rei, Thompson, Kocmi, Blain, Deutsch, Stewart, Zerva, Castilho, Lavie, and Foster}]{freitag2023results}
Markus Freitag, Nitika Mathur, Chi-kiu Lo, Eleftherios Avramidis, Ricardo Rei, Brian Thompson, Tom Kocmi, Frederic Blain, Daniel Deutsch, Craig Stewart, Chrysoula Zerva, Sheila Castilho, Alon Lavie, and George Foster. 2023.
\newblock \href {https://doi.org/10.18653/v1/2023.wmt-1.51} {Results of {WMT}23 metrics shared task: Metrics might be guilty but references are not innocent}.
\newblock In \emph{Proceedings of the Eighth Conference on Machine Translation}, pages 578--628, Singapore. Association for Computational Linguistics.

\bibitem[{Freitag et~al.(2022)Freitag, Rei, Mathur, Lo, Stewart, Avramidis, Kocmi, Foster, Lavie, and Martins}]{freitag2022results}
Markus Freitag, Ricardo Rei, Nitika Mathur, Chi-kiu Lo, Craig Stewart, Eleftherios Avramidis, Tom Kocmi, George Foster, Alon Lavie, and Andr{\'e} F.~T. Martins. 2022.
\newblock \href {https://aclanthology.org/2022.wmt-1.2/} {Results of {WMT}22 metrics shared task: Stop using {BLEU} -- neural metrics are better and more robust}.
\newblock In \emph{Proceedings of the Seventh Conference on Machine Translation (WMT)}, pages 46--68, Abu Dhabi, United Arab Emirates (Hybrid). Association for Computational Linguistics.

\bibitem[{Fujii et~al.(2024)Fujii, Nakamura, Loem, Iida, Ohi, Hattori, Shota, Mizuki, Yokota, and Okazaki}]{fujiicontinual}
Kazuki Fujii, Taishi Nakamura, Mengsay Loem, Hiroki Iida, Masanari Ohi, Kakeru Hattori, Hirai Shota, Sakae Mizuki, Rio Yokota, and Naoaki Okazaki. 2024.
\newblock Continual pre-training for cross-lingual llm adaptation: Enhancing japanese language capabilities.
\newblock In \emph{First Conference on Language Modeling}.

\bibitem[{Goyal et~al.(2021)Goyal, Gao, Chaudhary, Chen, Wenzek, Ju, Krishnan, Ranzato, Guzm{\'a}n, and Fan}]{Goyal2021TheFE}
Naman Goyal, Cynthia Gao, Vishrav Chaudhary, Peng-Jen Chen, Guillaume Wenzek, Da~Ju, Sanjan Krishnan, Marc'Aurelio Ranzato, Francisco Guzm{\'a}n, and Angela Fan. 2021.
\newblock \href {https://api.semanticscholar.org/CorpusID:235358129} {The flores-101 evaluation benchmark for low-resource and multilingual machine translation}.
\newblock \emph{Transactions of the Association for Computational Linguistics}, 10:522--538.

\bibitem[{Goyal et~al.(2022)Goyal, Gao, Chaudhary, Chen, Wenzek, Ju, Krishnan, Ranzato, Guzm{\'a}n, and Fan}]{goyal2022flores}
Naman Goyal, Cynthia Gao, Vishrav Chaudhary, Peng-Jen Chen, Guillaume Wenzek, Da~Ju, Sanjana Krishnan, Marc’Aurelio Ranzato, Francisco Guzm{\'a}n, and Angela Fan. 2022.
\newblock The flores-101 evaluation benchmark for low-resource and multilingual machine translation.
\newblock \emph{Transactions of the Association for Computational Linguistics}, 10:522--538.

\bibitem[{Guyot(2018)}]{guyot2018translating}
Alexandra Guyot. 2018.
\newblock Translating mircea eliade's" ivan" from romanian to english: A triangular approach using the french translation.

\bibitem[{He et~al.(2025)He, Wang, Xiong, and Liu}]{he2025box}
Jie He, Tao Wang, Deyi Xiong, and Qun Liu. 2025.
\newblock The box is in the pen: Evaluating commonsense reasoning in neural machine translation.
\newblock \emph{arXiv preprint arXiv:2503.03308}.

\bibitem[{Jaech et~al.(2024)Jaech, Kalai, Lerer, Richardson, El-Kishky, Low, Helyar, Madry, Beutel, Carney et~al.}]{jaech2024openai}
Aaron Jaech, Adam Kalai, Adam Lerer, Adam Richardson, Ahmed El-Kishky, Aiden Low, Alec Helyar, Aleksander Madry, Alex Beutel, Alex Carney, et~al. 2024.
\newblock Openai o1 system card.
\newblock \emph{arXiv preprint arXiv:2412.16720}.

\bibitem[{Jiao et~al.(2024)Jiao, Peng, Zong, Zhang, and Li}]{jiao2024gradable}
Hui Jiao, Bei Peng, Lu~Zong, Xiaojun Zhang, and Xinwei Li. 2024.
\newblock Gradable chatgpt translation evaluation.
\newblock \emph{Procesamiento del Lenguaje Natural}, 72:73--85.

\bibitem[{Jiao et~al.(2023{\natexlab{a}})Jiao, tse Huang, Wang, He, Liang, Wang, Shi, and Tu}]{jiao2023parrot}
Wenxiang Jiao, Jen tse Huang, Wenxuan Wang, Zhiwei He, Tian Liang, Xing Wang, Shuming Shi, and Zhaopeng Tu. 2023{\natexlab{a}}.
\newblock Parrot: Translating during chat using large language models tuned with human translation and feedback.
\newblock In \emph{Findings of EMNLP}.

\bibitem[{Jiao et~al.(2023{\natexlab{b}})Jiao, Wang, Huang, Wang, and Tu}]{jiao2023chatgpt}
Wenxiang Jiao, Wenxuan Wang, Jen-tse Huang, Xing Wang, and Zhaopeng Tu. 2023{\natexlab{b}}.
\newblock Is chatgpt a good translator? a preliminary study.
\newblock \emph{arXiv preprint arXiv:2301.08745}.

\bibitem[{Ko et~al.(2025)Ko, Son, and Choi}]{ko2025understand}
Hyunwoo Ko, Guijin Son, and Dasol Choi. 2025.
\newblock Understand, solve and translate: Bridging the multilingual mathematical reasoning gap.
\newblock \emph{arXiv preprint arXiv:2501.02448}.

\bibitem[{Kocmi et~al.(2024{\natexlab{a}})Kocmi, Avramidis, Bawden, Bojar, Dvorkovich, Federmann, Fishel, Freitag, Gowda, Grundkiewicz et~al.}]{kocmi2024findings}
Tom Kocmi, Eleftherios Avramidis, Rachel Bawden, Ond{\v{r}}ej Bojar, Anton Dvorkovich, Christian Federmann, Mark Fishel, Markus Freitag, Thamme Gowda, Roman Grundkiewicz, et~al. 2024{\natexlab{a}}.
\newblock Findings of the wmt24 general machine translation shared task: The llm era is here but mt is not solved yet.
\newblock In \emph{Proceedings of the Ninth Conference on Machine Translation}, pages 1--46.

\bibitem[{Kocmi et~al.(2024{\natexlab{b}})Kocmi, Avramidis, Bawden, Bojar, Dvorkovich, Federmann, Fishel, Freitag, Gowda, Grundkiewicz et~al.}]{kocmi2024preliminary}
Tom Kocmi, Eleftherios Avramidis, Rachel Bawden, Ondrej Bojar, Anton Dvorkovich, Christian Federmann, Mark Fishel, Markus Freitag, Thamme Gowda, Roman Grundkiewicz, et~al. 2024{\natexlab{b}}.
\newblock Preliminary wmt24 ranking of general mt systems and llms.
\newblock \emph{arXiv preprint arXiv:2407.19884}.

\bibitem[{Kong et~al.(2023)Kong, Zhao, Chen, Li, Qin, Sun, Zhou, Wang, and Dong}]{kong2023better}
Aobo Kong, Shiwan Zhao, Hao Chen, Qicheng Li, Yong Qin, Ruiqi Sun, Xin Zhou, Enzhi Wang, and Xiaohang Dong. 2023.
\newblock Better zero-shot reasoning with role-play prompting.
\newblock \emph{arXiv preprint arXiv:2308.07702}.

\bibitem[{Nida(1964)}]{nida1964toward}
Eugene~Albert Nida. 1964.
\newblock \emph{Toward a science of translating: with special reference to principles and procedures involved in Bible translating}.
\newblock Brill Archive.

\bibitem[{Ordudari(2007)}]{ordudari2007translation}
Mahmoud Ordudari. 2007.
\newblock Translation procedures, strategies and methods.
\newblock \emph{Translation journal}, 11(3):8.

\bibitem[{Ouyang et~al.(2022)Ouyang, Wu, Jiang, Almeida, Wainwright, Mishkin, Zhang, Agarwal, Slama, Ray et~al.}]{ouyang2022training}
Long Ouyang, Jeffrey Wu, Xu~Jiang, Diogo Almeida, Carroll Wainwright, Pamela Mishkin, Chong Zhang, Sandhini Agarwal, Katarina Slama, Alex Ray, et~al. 2022.
\newblock Training language models to follow instructions with human feedback.
\newblock \emph{Advances in neural information processing systems}, 35:27730--27744.

\bibitem[{Peng et~al.(2023)Peng, Ding, Zhong, Shen, Liu, Zhang, Ouyang, and Tao}]{peng2023towards}
Keqin Peng, Liang Ding, Qihuang Zhong, Li~Shen, Xuebo Liu, Min Zhang, Yuanxin Ouyang, and Dacheng Tao. 2023.
\newblock Towards making the most of chatgpt for machine translation.
\newblock \emph{arXiv preprint arXiv:2303.13780}.

\bibitem[{Qin et~al.(2024)Qin, Li, Zou, Liu, Xia, Huang, Ye, Yuan, Liu, Li et~al.}]{qin2024o1}
Yiwei Qin, Xuefeng Li, Haoyang Zou, Yixiu Liu, Shijie Xia, Zhen Huang, Yixin Ye, Weizhe Yuan, Hector Liu, Yuanzhi Li, et~al. 2024.
\newblock O1 replication journey: A strategic progress report--part 1.
\newblock \emph{arXiv preprint arXiv:2410.18982}.

\bibitem[{Rei et~al.(2020)Rei, Stewart, Farinha, and Lavie}]{rei-etal-2020-comet}
Ricardo Rei, Craig Stewart, Ana~C Farinha, and Alon Lavie. 2020.
\newblock \href {https://doi.org/10.18653/v1/2020.emnlp-main.213} {{COMET}: A neural framework for {MT} evaluation}.
\newblock In \emph{Proceedings of the 2020 Conference on Empirical Methods in Natural Language Processing (EMNLP)}, pages 2685--2702, Online. Association for Computational Linguistics.

\bibitem[{Ringmar(2012)}]{ringmar2012relay}
Martin Ringmar. 2012.
\newblock Relay translation.
\newblock \emph{Handbook of translation studies}, 3:141--144.

\bibitem[{Schulman et~al.(2017)Schulman, Wolski, Dhariwal, Radford, and Klimov}]{schulman2017proximal}
John Schulman, Filip Wolski, Prafulla Dhariwal, Alec Radford, and Oleg Klimov. 2017.
\newblock Proximal policy optimization algorithms.
\newblock \emph{arXiv preprint arXiv:1707.06347}.

\bibitem[{Shao et~al.(2024)Shao, Wang, Zhu, Xu, Song, Bi, Zhang, Zhang, Li, Wu et~al.}]{shao2024deepseekmath}
Zhihong Shao, Peiyi Wang, Qihao Zhu, Runxin Xu, Junxiao Song, Xiao Bi, Haowei Zhang, Mingchuan Zhang, YK~Li, Y~Wu, et~al. 2024.
\newblock Deepseekmath: Pushing the limits of mathematical reasoning in open language models.
\newblock \emph{arXiv preprint arXiv:2402.03300}.

\bibitem[{Shen et~al.(2025)Shen, Zhang, Zhang, Xu, and Zhao}]{shen2025vlmr1}
Haozhan Shen, Zilun Zhang, Qianqian Zhang, Ruochen Xu, and Tiancheng Zhao. 2025.
\newblock Vlm-r1: A stable and generalizable r1-style large vision-language model.
\newblock \url{https://github.com/om-ai-lab/VLM-R1}.
\newblock Accessed: 2025-02-15.

\bibitem[{Ul~Hassan et~al.(2024)Ul~Hassan, Yu, Wang, Li, Gao, Yang, and Mao}]{ul2024lkmt}
Muhammad~Naeem Ul~Hassan, Zhengtao Yu, Jian Wang, Ying Li, Shengxiang Gao, Shuwan Yang, and Cunli Mao. 2024.
\newblock Lkmt: Linguistics knowledge-driven multi-task neural machine translation for urdu and english.
\newblock \emph{Computers, Materials \& Continua}, 81(1).

\bibitem[{Wang et~al.(2024{\natexlab{a}})Wang, Meng, Liang, and Zhou}]{jiaan2024drt}
Jiaan Wang, Fandong Meng, Yunlong Liang, and Jie Zhou. 2024{\natexlab{a}}.
\newblock Drt-o1: Optimized deep reasoning translation via long chain-of-thought.
\newblock \emph{arXiv preprint arXiv:2412.17498}.

\bibitem[{Wang et~al.(2024{\natexlab{b}})Wang, Meng, Zhang, and Zhou}]{wang2024retrieval}
Jiaan Wang, Fandong Meng, Yingxue Zhang, and Jie Zhou. 2024{\natexlab{b}}.
\newblock Retrieval-augmented machine translation with unstructured knowledge.
\newblock \emph{arXiv preprint arXiv:2412.04342}.

\bibitem[{Wang et~al.(2024{\natexlab{c}})Wang, Zeng, Liu, Meng, Zhou, and Zhang}]{wang-etal-2024-taste}
Yutong Wang, Jiali Zeng, Xuebo Liu, Fandong Meng, Jie Zhou, and Min Zhang. 2024{\natexlab{c}}.
\newblock \href {https://doi.org/10.18653/v1/2024.acl-long.333} {{T}as{T}e: Teaching large language models to translate through self-reflection}.
\newblock In \emph{Proceedings of the 62nd Annual Meeting of the Association for Computational Linguistics (Volume 1: Long Papers)}, pages 6144--6158, Bangkok, Thailand. Association for Computational Linguistics.

\bibitem[{Wei et~al.(2022)Wei, Wang, Schuurmans et~al.}]{wei2022chain}
Jason Wei, Xuezhi Wang, Dale Schuurmans, et~al. 2022.
\newblock Chain of thought prompting elicits reasoning in large language models.
\newblock In \emph{Advances in Neural Information Processing Systems (NeurIPS)}.

\bibitem[{Wu et~al.(2024)Wu, Vu, Qu, Foster, and Haffari}]{wu2024adapting}
Minghao Wu, Thuy-Trang Vu, Lizhen Qu, George Foster, and Gholamreza Haffari. 2024.
\newblock Adapting large language models for document-level machine translation.
\newblock \emph{arXiv preprint arXiv:2401.06468}.

\bibitem[{Xu et~al.(2024{\natexlab{a}})Xu, Kim, Sharaf, and Awadalla}]{xuparadigm}
Haoran Xu, Young~Jin Kim, Amr Sharaf, and Hany~Hassan Awadalla. 2024{\natexlab{a}}.
\newblock A paradigm shift in machine translation: Boosting translation performance of large language models.
\newblock In \emph{The Twelfth International Conference on Learning Representations}.

\bibitem[{Xu et~al.(2024{\natexlab{b}})Xu, Murray, Koehn, Hoang, Eriguchi, and Khayrallah}]{xu2024x}
Haoran Xu, Kenton Murray, Philipp Koehn, Hieu Hoang, Akiko Eriguchi, and Huda Khayrallah. 2024{\natexlab{b}}.
\newblock X-alma: Plug \& play modules and adaptive rejection for quality translation at scale.
\newblock \emph{arXiv preprint arXiv:2410.03115}.

\bibitem[{Yang et~al.(2024)Yang, Yang, Zhang, Hui, Zheng, Yu, Li, Liu, Huang, Dong, Wei, Lin, Yang, Tu, Zhang, Yang, Yang, Zhou, Lin, Dang, Lu, Bao, Yang, Yu, Li, Xue, Zhang, Zhu, Men, Lin, Li, Xia, Ren, Ren, Fan, Su, Zhang, Wan, Liu, Cui, Zhang, Qiu, Quan, and Wang}]{Yang2024Qwen25TR}
Qwen~An Yang, Baosong Yang, Beichen Zhang, Binyuan Hui, Bo~Zheng, Bowen Yu, Chengyuan Li, Dayiheng Liu, Fei Huang, Guanting Dong, Haoran Wei, Huan Lin, Jian Yang, Jianhong Tu, Jianwei Zhang, Jianxin Yang, Jiaxin Yang, Jingren Zhou, Junyang Lin, Kai Dang, Keming Lu, Keqin Bao, Kexin Yang, Le~Yu, Mei Li, Mingfeng Xue, Pei Zhang, Qin Zhu, Rui Men, Runji Lin, Tianhao Li, Tingyu Xia, Xingzhang Ren, Xuancheng Ren, Yang Fan, Yang Su, Yi-Chao Zhang, Yunyang Wan, Yuqi Liu, Zeyu Cui, Zhenru Zhang, Zihan Qiu, Shanghaoran Quan, and Zekun Wang. 2024.
\newblock \href {https://api.semanticscholar.org/CorpusID:274859421} {Qwen2.5 technical report}.
\newblock \emph{ArXiv}, abs/2412.15115.

\bibitem[{Yao et~al.(2023)Yao, Jiang, Bobinac, Yang, and Hu}]{yao2023benchmarking}
Binwei Yao, Ming Jiang, Tara Bobinac, Diyi Yang, and Junjie Hu. 2023.
\newblock Benchmarking machine translation with cultural awareness.
\newblock \emph{arXiv preprint arXiv:2305.14328}.

\bibitem[{Zhang et~al.(2023{\natexlab{a}})Zhang, Wang, Qin, Shi, Wang, and Chen}]{zhang2023understanding}
Huaao Zhang, Qiang Wang, Bo~Qin, Zelin Shi, Haibo Wang, and Ming Chen. 2023{\natexlab{a}}.
\newblock Understanding and improving the robustness of terminology constraints in neural machine translation.
\newblock In \emph{Proceedings of the 61st Annual Meeting of the Association for Computational Linguistics (Volume 1: Long Papers)}, pages 6029--6042.

\bibitem[{Zhang et~al.(2023{\natexlab{b}})Zhang, Rajabi, Duh, and Koehn}]{zhang2023machine}
Xuan Zhang, Navid Rajabi, Kevin Duh, and Philipp Koehn. 2023{\natexlab{b}}.
\newblock Machine translation with large language models: Prompting, few-shot learning, and fine-tuning with qlora.
\newblock In \emph{Proceedings of the Eighth Conference on Machine Translation}, pages 468--481.

\bibitem[{Zhang et~al.(2024)Zhang, Wu, Yang, Shu, Xiao, Kong, and Sang}]{zhang2024o1}
Yuxiang Zhang, Shangxi Wu, Yuqi Yang, Jiangming Shu, Jinlin Xiao, Chao Kong, and Jitao Sang. 2024.
\newblock o1-coder: an o1 replication for coding.
\newblock \emph{arXiv preprint arXiv:2412.00154}.

\bibitem[{Zhao et~al.(2024)Zhao, Yin, Zeng, Wang, Shi, Lyu, Wang, Luo, and Zhang}]{zhao2024marco}
Yu~Zhao, Huifeng Yin, Bo~Zeng, Hao Wang, Tianqi Shi, Chenyang Lyu, Longyue Wang, Weihua Luo, and Kaifu Zhang. 2024.
\newblock Marco-o1: Towards open reasoning models for open-ended solutions.
\newblock \emph{arXiv preprint arXiv:2411.14405}.

\bibitem[{Zhu et~al.(2025)Zhu, Pan, Jian, and Xiong}]{zhu2025overcoming}
Shaolin Zhu, Leiyu Pan, Dong Jian, and Deyi Xiong. 2025.
\newblock Overcoming language barriers via machine translation with sparse mixture-of-experts fusion of large language models.
\newblock \emph{Information Processing \& Management}, 62(3):104078.

\bibitem[{Zouhar et~al.(2024)Zouhar, Chen, Lam, Moghe, and Haddow}]{zouhar2024pitfalls}
Vil{\'e}m Zouhar, Pinzhen Chen, Tsz~Kin Lam, Nikita Moghe, and Barry Haddow. 2024.
\newblock \href {https://doi.org/10.18653/v1/2024.wmt-1.121} {Pitfalls and outlooks in using {COMET}}.
\newblock In \emph{Proceedings of the Ninth Conference on Machine Translation}, pages 1272--1288, Miami, Florida, USA. Association for Computational Linguistics.

\end{thebibliography}

\appendix

\newpage

\section{The real-world MT dataset source}\label{sec:appendix A}
All sources of our parallel corpus come from following:

(1) \url{https://machinetranslate.org/wmt}

(2) \url{https://www.un.org/dgacm/zh/content/uncorpus/}

(3) \url{https://huggingface.co/datasets/joefox/newstest-2017-2019-ru_zh/tree/main}

(4) \url{https://www.jizhi-dataset.top/index/category/detail/15}

\section{Human translation CoT templates}\label{sec:appendix B}
\begin{mybox}
[Hierarchical Translation]
<think>

1. Analyze the sentence structure and identify the core elements (subject, verb, object).

2. Translate the sentence from the origin language to the target language, focusing on the core elements.

3. Review the translation for basic accuracy and grammatical structure.

4. Identify areas that need further refinement (e.g., word choice, tense, or word order).

5. Modify the translation to improve fluency and coherence, considering the context.

6. Finalize the translation by ensuring it retains the original meaning while improving readability. 

</think>
\end{mybox}

\begin{mybox}
[Triangulating Translation]
<think>

1. Identify basic elements: Break down the sentence into its main components and identify the key subject, verb, and object.

2. Translate to intermediate language: Convert these elements into an intermediate language structure (e.g., simple syntactic rules or function names).

3. Refine back to target language: Translate from the intermediate language back to the target language, adjusting for syntactic norms and idiomatic expressions.

4. Check for accuracy: Ensure that the meaning is preserved in the translated sentence by checking noun-verb agreement and connectors.

5. Adjust word order: Modify word order to ensure that it aligns with the target language’s grammatical structure.

6. Final refinement: Review the translation for naturalness, idiomatic use, and overall flow.

</think>
\end{mybox}

\begin{mybox}
[Back Translation]
<think>

1. Analyze the provided context in the source language.

2. Translate the source text to the target language.

3. Perform back translation from the target language to the source language.

4. Compare the back translation with the original source context.

5. Evaluate whether the meaning of the back translation aligns with the original.

6. If discrepancies are identified, adjust the target language translation to enhance consistency with the original meaning.

7. Finalize the translation by ensuring both forward and back translations accurately align across all languages involved.

</think>
\end{mybox}

\begin{mybox}
[Context-aware Translation]
<think>

1. Analyze the current sentence, along with the previous sentences, to understand the overall conversation context.

2. Identify key elements like tone, formality, or subject matter based on the ongoing conversation.

3. Translate the sentence while ensuring that the translation is aligned with the tone, style, and subject of the preceding dialogue.

4. If any ambiguity exists in the translation due to context, refine the translation to better fit the conversation flow.

5. Verify that the translation maintains coherence with the larger conversation, ensuring consistency in language and tone.

6. Finalize the translation by cross-checking it with the conversation’s context to ensure it feels natural and appropriately aligned. 

</think>
\end{mybox}

\begin{mybox}
[Translation Explanation]
<think>

1. Analyze the source sentence and identify the key elements (verbs, subjects, objects, etc.).

2. Based on these elements, determine the most suitable translation strategy (literal vs. idiomatic).

3. Select the best translation for each word or phrase, considering context and language-specific structures.

4. Explain the rationale behind choosing specific words or phrases.

5. After completing the initial translation, review each translation decision and explain any adjustments made for fluency or accuracy.

6. Provide a final explanation for the translation choices, discussing any trade-offs made between literal meaning and contextual appropriateness.

</think>
\end{mybox}

\begin{mybox}
[Structural Transformation]
<think>

1. Analyze the sentence's syntactic structure in the source language (e.g., identify whether it’s active or passive).

2. Determine the most appropriate syntactic structure in the target language (e.g., whether it needs to be rephrased from active to passive or vice versa).

3. Adjust the word order and grammatical structure in the target language to match the sentence’s meaning, while maintaining clarity.

4. Translate the sentence, ensuring that subject-verb-object relationships and other syntactic elements align with target language norms.

5. After the translation, check the sentence's grammar and overall flow in the target language, making sure it is clear and fluid.

6. If the sentence feels awkward or unnatural, refine the structure by adjusting word choice or reordering components. 

</think>
\end{mybox}

\section{Prompt for CoT instance Construction}\label{sec:appendix C}

\subsection{Incorporating our human CoTs into Parallel Corpus}
This prompt is used to construct cold start data for R1-T1, with human-constructed CoT templates as guidance.
\begin{mybox}
Assume that you are a student engaged in translating a sentence from \verb|{source_lang}| to \verb|{target_lang}|.
Now you have both the source sentence and the target sentence, and need to analyze how to translate from the source sentence to the given target sentence based on the provided Thinking Chain Guide. And output the chain-of-thought trajectory from source to target sentence.\\

The \verb|{source_lang}| statement is as follows:

\verb|<Source Sentence>|

\verb|{source_text}|

\verb|</Source Sentence>| \\

The \verb|{target_lang}| statement is as follows:

\verb|<Target Sentence>|

\verb|{target_text}|

\verb|</Target Sentence>| \\

You continuously reflect on how to translate the source sentence to the given target sentence based on the thinking guidance provided. \\ 

The given Thinking Chain Guide is as follows:

\verb|<Thinking Chain Guide>|

\verb|{thought_chains}|

\verb|</Thinking Chain Guide>| \\

Please refine the entire analysis process into a complete self-reflective description (in the present tense). For self-reflection, you can refer to the following thinking steps: directly output the self-reflective description in the \verb|<think></think>| tags, without any additional descriptions or explanations. Each line in the reflective description can be viewed as a reasoning step in the translation process.
\end{mybox}

\subsection{Incorporating LLM generated CoTs into Parallel Corpus}

This prompt is used to construct cold start data for variant (6) \textit{Q2.5-7b (w/o human CoT) + SFT + RL} in ablation study Section \ref{sec:ablation}: Training with the same setting as R1-T1, but replacing the human CoT with a purely LLM-generated CoT dataset for cold start. The key difference lies in whether there is a Thinking Chain Guide part in the two prompts.

\begin{mybox}
\noindent
Assume that you are a linguist engaged in translating a sentence from \verb|{source_lang}| to \verb|{target_lang}|.\\
Now you have both the source sentence and the target sentence, and you need to imitate the daily translation thought process of a human translator, analyzing how to translate from the source sentence to the given target sentence. And output the chain-of-thought trajectory from source to target sentence.\\

The \verb|{source_lang}| statement is as follows:

\verb|<Source Sentence>|\\
\verb|{source_text}|\\
\verb|</Source Sentence>|\\

The \verb|{target_lang}| statement is as follows:

\verb|<Target Sentence>|\\
\verb|{target_text}|\\
\verb|</Target Sentence>|\\

During the analysis process, you continually reflect on how to translate from the source sentence to the given target sentence.\\
Please refine the entire analysis process into a complete self-reflective description (in the present tense). For self-reflection, you can refer to the following thinking steps: directly output the self-reflective description in the \verb|<think></think>| tags, without any additional descriptions or explanations. Each line in the reflective description can be viewed as a reasoning step in the translation process.
\end{mybox}

\section{Human Evaluation Comments}\label{sec:appendix E}

To further interpret the human evaluation scores in Table~\ref{tab:human_eval}, we present summarized comments from professional translators across 12 languages in Table~\ref{tab:human_eval_comments}. The observations highlight comparative strengths and weaknesses of R1-T1 and the baseline in different directions.

\begin{table*}[ht]
\centering
\small
\begin{tabular}{p{2.2cm} p{5.7cm} p{5.7cm}}
\toprule
\textbf{Language} & \textbf{EN/ZH $\rightarrow$ XX} & \textbf{XX $\rightarrow$ EN/ZH} \\
\midrule
Arabic & R1-T1 shows more accurate lexical choices and clearer sentence structure. & R1-T1 produces more fluent and accurate translations, especially in terminology. \\
German & R1-T1 uses more consistent wording and has clearer structure, though both systems show grammar issues. & R1-T1 offers better logic and coherence, especially in conjunctions and clause reordering, but can be verbose. \\
Greek & R1-T1 better translates proper nouns, verbs, and terms. & R1-T1 follows source logic more faithfully, though both systems exhibit omission/addition issues. \\
Spanish & R1-T1 has fewer redundant expressions and better fidelity to the source. & R1-T1 is more accurate and natural, especially in domain terms; Result 2 has omissions and mistranslations. \\
French & R1-T1 generally more natural and faithful, though sometimes has term issues. & Mixed: R1-T1 better in culture/biomedical; Result 2 slightly better on Flores subset. \\
Italian & Result 2 slightly better in en$\rightarrow$it; closer performance in cn$\rightarrow$it. & R1-T1 more fluent in it$\rightarrow$en; Result 2 better in it$\rightarrow$zh. \\
Japanese & Both systems weak on grammar; Result 2 slightly stronger. & Result 1 follows source order; Result 2 better with restructuring and idioms. \\
Korean & Similar quality overall; Result 2 has better phrasing but more hallucinations. & Similar overall; Result 1 slightly clearer in logic. \\
Malay & Result 2 better with term precision and clarity. & R1-T1 more accurate and coherent overall. \\
Polish & Result 1 avoids hallucination but invents words; Result 2 more stable. & Result 1 better overall, fewer critical errors. \\
Portuguese & Similar quality; R1-T1 slightly more accurate. & R1-T1 handles fluency better in zh$\rightarrow$pt. \\
Russian & R1-T1 preserves source structure better and uses terms more precisely. & Similar performance; R1-T1 generally more faithful, Result 2 more fluent at times. \\
\bottomrule
\end{tabular}
\caption{Summary of human translator comments comparing R1-T1 and baseline across translation directions and languages. Here Result 2 refers to Q2.5-7B-SFT in Table \ref{tab:human_eval}. EN/ZH $\rightarrow$ XX means translation directions from English or Chinese to target languages, and vice versa.}
\label{tab:human_eval_comments}
\end{table*}

\section{CoT Self-evolution Analysis}\label{sec:appendix F}
To provide a more intuitive evaluation of the CoT self-evolution mechanism, we conduct a visualization for the translation of a complex sentence from Chinese to English, with and without the integration of self-evolving CoTs in Fig. \ref{fig1}. 
In both instances, the translated sentence correctly conveys the meaning, but the version utilizing self-evolution provides more nuanced translations. The description without CoTs includes generic phrasing, such as "the meeting was held," whereas the self-evolved CoT version includes additional context, such as "the critical meeting was held to address urgent concerns."

Through this analysis, it becomes evident that the self-evolving CoT improves the model's ability to adapt over time, refining its translation strategies. The self-evolving CoT allows the model to generate more diverse and accurate translations by incorporating expert-level reasoning strategies and incorporating feedback from prior outputs. This demonstrates that CoT self-evolution is effective in enhancing the model's reliability and its capacity to provide richer, context-aware translations across languages.

\section{Evaluating R1-T1 on Additional Languages}

\begin{table*}[t]
    \centering
    \resizebox{\linewidth}{!}{
        \begin{tabular}{l@{\hskip 0.05in}c@{\hskip 0.05in}c@{\hskip 0.05in}c@{\hskip 0.05in}c@{\hskip 0.05in}c@{\hskip 0.05in}c@{\hskip 0.05in}c@{\hskip 0.05in}c@{\hskip 0.05in}c@{\hskip 0.05in}c@{\hskip 0.05in}c@{\hskip 0.05in}c@{\hskip 0.05in}c@{\hskip 0.05in}c@{\hskip 0.05in}c@{\hskip 0.05in}c@{\hskip 0.05in}c@{\hskip 0.05in}c@{\hskip 0.05in}c@{\hskip 0.05in}c@{\hskip 0.05in}c}
            \toprule
    \multirow{2}{*}{\textbf{Models}} & \multicolumn{5}{c}{\hspace{-1.5em}\textbf{xx2en}}       & \multicolumn{5}{c}{\hspace{-0.2em}\textbf{en2xx}} & \multicolumn{5}{c}{\hspace{-0.2em}\textbf{xx2zh}} & \multicolumn{5}{c}{\hspace{-0.2em}\textbf{zh2xx}} & \textbf{avg}\\
    \cmidrule(l{0.3em}r{1em}){2-6} \cmidrule(l{0.3em}r{1em}){7-11} \cmidrule(l{0.3em}r{1em}){12-16} \cmidrule(l{0.3em}r{1em}){17-21}
        & \textbf{pt} & \textbf{id} & \textbf{es} & \textbf{it} & \textbf{ko} & \textbf{pt} & \textbf{id} & \textbf{es} & \textbf{it} & \textbf{ko} & \textbf{pt} & \textbf{id} & \textbf{es} & \textbf{it} & \textbf{ko} & \textbf{pt} & \textbf{id} & \textbf{es} & \textbf{it} & \textbf{ko} & \textbf{-} \\ \midrule  
        \multicolumn{21}{c}{\textbf{Existing LLMs}} \\
    \midrule
    Q2.5-7b & \textbf{0.830} & 0.787 & 0.721 & 0.732 & 0.663 & 0.845 & 0.733 & 0.676 & 0.649 & 0.258 & 0.555 & 0.560 & 0.539 & 0.558 & 0.558 & 0.640 & 0.577 &0.544 & 0.530 & 0.343 & 0.615 \\
    Q2.5-7b-Ins & 0.711 & 0.541 & 0.642 & 0.642 & 0.664 & 0.703 & 0.574 & 0.580 & 0.503 & 0.056 & 0.494 & 0.500 & 0.453 & 0.276 & 0.548 & 0.320 & 0.319 & 0.332 & 0.260 & 0.047 & 0.458 \\
    DS-R1-D & 0.709 & 0.545 & 0.586 & 0.580 & 0.271 & 0.068 & -0.148 & -0.078 & -0.295 & -1.275 & 0.250 & 0.111 & 0.198 & 0.188 & -0.036 & -0.228 & -0.302 & -0.218 & -0.397 & -0.950 & -0.021 \\
    DS-R1-DM & 0.690 & 0.486 & 0.588 & 0.576 & 0.280 & 0.174 & -0.011 & 0.098 & -0.140 & -0.831 & 0.272 & 0.055 & 0.225 & 0.154 & -0.166 & -0.181 & -0.345 & -0.155 & -0.341 & -0.781 & -0.032 \\
    \midrule
    \multicolumn{21}{c}{\textbf{Ablation Groups}} \\
    \midrule
    Q2.5-7b + SFT  & 0.829 & 0.778 & \textbf{0.732} & \textbf{0.743} & 0.661 & 0.831 & 0.727 & 0.685 & 0.655 & 0.406 & 0.591 & 0.580 & 0.561 & 0.577 & 0.540 & 0.612 & 0.558 & 0.534 & 0.485 & 0.371 & 0.623 \\
    Q2.5-7b + RL  & 0.826 & 0.785 & 0.722 & 0.732 & 0.666 & 0.826 & 0.680 & 0.685 & 0.652 & 0.533 & 0.590 & 0.599 & 0.567 & 0.589 & 0.572 & 0.621 & 0.534 & 0.547 & 0.497 & 0.446 & 0.633 \\\midrule
    \textbf{R1-T1} & 0.829 & \textbf{0.789} & 0.723 & 0.738 & \textbf{0.672} & \textbf{0.854} & \textbf{0.742} & \textbf{0.708} & \textbf{0.690} & \textbf{0.566} & 0.609 & \textbf{0.608} & \textbf{0.574} & \textbf{0.591} & \textbf{0.597} & \textbf{0.654} & \textbf{0.592} & \textbf{0.587} & \textbf{0.561} & \textbf{0.525} & \textbf{0.660} \\
    \bottomrule
    \end{tabular}
    }
    \caption{Translation performance of R1-T1 on five additional languages. \textbf{pt}, \textbf{id}, \textbf{es}, \textbf{it}, \textbf{ko} mean Portuguese, Indonesian, Spanish, Italian, Korean. Other notations are the same as Table~\ref{tab:unseen_flores} and Table~\ref{tab:ablation_study}.}
    \label{tab:new_five_flores}
\end{table*}

As shown in Table~\ref{tab:new_five_flores}, we evaluated R1-T1 on five additional languages to further examine its generalization abilities, \textbf{which brings the total number of language studied in this paper up to 16 and number of translation directions up to 60}. By comparing R1-T1 with existing LLMs and methods in the ablation groups, we can see the strong average translation performances of our approach on these additional languages, indicating a promising potential for R1-T1 to be applied in broader language scenarios.

\end{document}